\documentclass{article}

\usepackage{neurips_2024}

\usepackage[utf8]{inputenc} 
\usepackage[backend=bibtex]{biblatex}
\usepackage[T1]{fontenc}    
\usepackage{hyperref}       
\usepackage{url}            
\usepackage{booktabs}       
\usepackage{amsfonts}       
\usepackage{nicefrac}       
\usepackage{microtype}      
\usepackage{xcolor}         
\usepackage{amsmath}
\usepackage{ragged2e}

\usepackage[none]{hyphenat} 
\usepackage{multirow}
\usepackage[normalem]{ulem}
\usepackage{bbding} 

\usepackage{multicol} 
\usepackage{array} 
\usepackage{threeparttable}
\usepackage{cutwin}
\usepackage{subcaption}
\usepackage{graphicx} 
\usepackage{wrapfig}
\usepackage{xcolor}
\usepackage{colortbl}

\justifying
\title{HDRGS: High Dynamic Range Gaussian Splatting}

\author{
  Jiahao Wu$^1$, Lu Xiao$^1$,Rui Peng$^1$, Kaiqiang Xiong$^1$, Ronggang Wang$^1$\thanks{Corresponding author }\\
  $^1$School of Electronic and Computer Engineering, Peking University \\
}

\begin{document}

\maketitle

\begin{abstract}
\label{ch:abstract}

Recent years have witnessed substantial advancements in the field of 3D reconstruction from 2D images, particularly following the introduction of the neural radiance field (NeRF) technique. However, reconstructing a 3D high dynamic range (HDR) radiance field, which aligns more closely with real-world conditions, from 2D multi-exposure low dynamic range (LDR) images continues to pose significant challenges. Approaches to this issue fall into two categories: grid-based and implicit-based. Implicit methods, using multi-layer perceptrons (MLP), face inefficiencies, limited solvability, and overfitting risks. Conversely, grid-based methods require significant memory and struggle with image quality and long training times. In this paper, we introduce Gaussian Splatting—a recent, high-quality, real-time 3D reconstruction technique—into this domain. We further develop the High Dynamic Range Gaussian Splatting (HDR-GS) method, designed to address the aforementioned challenges. This method enhances color dimensionality by including luminance and uses an asymmetric grid for tone-mapping, swiftly and precisely converting pixel irradiance to color. Our approach improves HDR scene recovery accuracy and integrates a novel coarse-to-fine strategy to speed up model convergence, enhancing robustness against sparse viewpoints and exposure extremes, and preventing local optima. Extensive testing confirms that our method surpasses current state-of-the-art techniques in both synthetic and real-world scenarios. Code will be released at \url{https://github.com/WuJH2001/HDRGS}

\end{abstract}

\section{Introduction}
\label{sec:Introduction}

In recent years, significant progress has been made in 3D reconstruction technology  \cite{yao2018mvsnet,mildenhall2021nerf,kerbl20233d}, but these technologies typically assume constant exposure conditions of input images, thus restoring scenes with low dynamic range (LDR). However, high dynamic range (HDR) scenes \cite{reinhard2020high, wang2023glowgan}, which are more consistent with the physical world, offer a broader dynamic range and provide a superior visual experience for humans. Traditional HDR image reconstruction techniques 
\cite{  mertens2009exposure,wang2022learning,debevec2023recovering} still focus on 2D images. How to reconstruct 3D HDR scenes from multi-exposure unstructured LDR images remains a question worthy of investigation. 

Current methodologies in this domain can be divided into two distinct categories: explicit-based, represented by HDR-plenoxel \cite{jun2022hdr}, and implicit-based, represented by HDR-NeRF \cite{huang2022hdr}. While these methods have achieved some impressive results, HDR-plenoxel relies on grids and spherical harmonics, complicating the construction of high-quality 3D HDR radiance fields and requiring substantial memory. HDR-NeRF models the entire HDR scene and Camera Response Function (CRF)\cite{dufaux2016high} using implicit MLPs, resulting in poor interpretability and slow training and rendering speeds, which poses challenges for real-world applications.With the recent introduction of 3D Gaussian Splatting technology \cite{kerbl20233d}, which can efficiently and effectively complete 3D reconstruction, it appears to offer a direction for addressing the aforementioned issues.

In this paper, we introduce a method based on 3D Gaussian Splatting (HDRGS) that can reconstruct a 3D HDR scene by simulating the complete physical imaging process of a camera, projecting radiance to pixel color. Each Gaussian point radiates radiance \(r\) (with potential values up to positive infinity). Through the splatting, we capture the pixel irradiance \(E\)  on the HDR image plane. The exposure from irradiance  \(E\) over a specific exposure time  \(t\) is then transformed through a series of functions to compute pixel values \(C\).  One critical aspect of our method involves modeling these function transformations, particularly addressing the uneven distribution of irradiance \(E\). We developed an uneven asymmetric grid \(g\) to facilitate a rapid and efficient mapping from irradiance \(E\)  to pixel value \(C\) , ensuring that the entire process remains differentiable and only requires LDR images for supervision to accurately construct the HDR radiance field. To mitigate the grid’s discrete nature and the tendency to fall into local optima, we implement a novel coarse-to-fine strategy. Initially, a sigmoid function serves as the tone mapper to initialize Gaussian points during the coarse phase. In the subsequent fine phase, we employ the asymmetric grid \(g\)  as the tone mapper, co-learning its parameters with the Gaussian points.

Our method’s effectiveness is demonstrated through comparative experiments with other approaches 
\cite{mildenhall2021nerf,martin2021nerf,jun2022hdr,huang2022hdr,kerbl20233d}  and ablation studies on our primary modules. The results indicate that our model surpasses current state-of-the-art methods. 
\begin{enumerate}
\item [1)] 
Our method can efficiently reconstructs HDR radiance fields from a series of LDR images with varied exposures within minutes, and it consistently surpasses current state-of-the-art methods across both synthetic and real datasets. 
\item [2)] 
We introduce a novel coarse-to-fine strategy that enhances the model's ability to reconstruct complex scenes, significantly improving both reconstruction speed and image quality.
\item [3)] 
We propose a differentiable, asymmetric grid to model the tone mapper, providing enhanced expressive power over symmetric grids.
\end{enumerate}

\begin{figure}[t]
    \centering
    \includegraphics[width=\textwidth,trim=2.8cm 3.3cm 3.3cm 3.3cm]{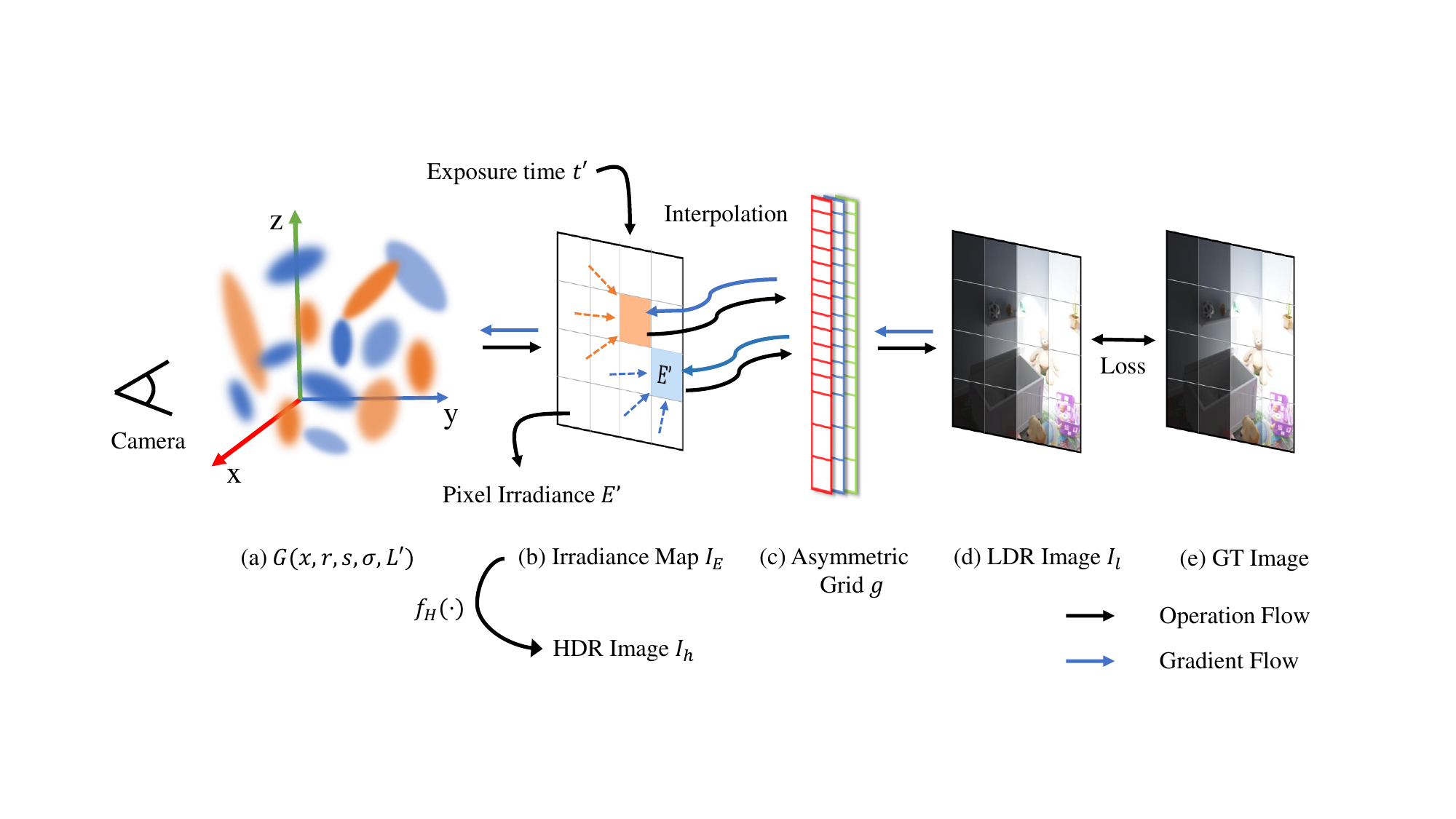}
    \caption{ \textbf{Illustration of HDRGS.} We redefine the color of Gaussian points as radiance \(L\), enabling it to meet the primary requirement for reconstructing the HDR radiance field. After splatting, pixel irradiance \(E'(\mathbf{p})\)  is obtained. Then, the differentiable asymmetric grid \(g\) maps the exposure under exposure time \( t \) to LDR pixel color. The HDR image \(I_h\) can be derived by applying the function \(f_H(\cdot)\) to each \(E'\), with detailed explanations provided in subsequent sections.}
    \label{fig:HDR-GS pipeline}
\end{figure}

\section{Related work}
\label{gen_inst}

\textbf{HDR Imaging.} The typical approach in traditional HDR imaging \cite{reinhard2020high} is crucial for enhancing people's immersive visual experiences. Currently, tone mapping, generating LDR from HDR, is a mature field \cite{reinhard2020high, wang2022learning}, but its inverse tone mapping (ITM) \cite{rempel2007ldr2hdr}, recovering HDR from LDR, still faces many challenges. The simplest method is multi-frame LDR image synthesis with different exposures to create an HDR image. As the name suggests, this involves extracting the most visible parts from each LDR image and combining them into a single HDR image. However, such methods based on multiple exposures \cite{debevec2023recovering} are typically constrained by the need for multiple exposure LDR images and are prone to artifacts caused by camera movements during LDR image capture. The emergence of deep learning in recent years has brought new solutions, with learning-based methods being categorized into indirect and direct approaches. Indirect methods \cite{endo2017deep, lee2018deep} typically predict multiple LDR images with different exposures, which are then merged into an HDR image (Indirect ITM), while direct methods \cite{eilertsen2017hdr, marnerides2018expandnet} involve neural networks directly predicting HDR images (Direct ITM). However, these methods can only synthesize HDR images with the original camera poses.

\textbf{Novel View Synthesis.} Previous methods formalized the light field \cite{levoy2023light} or Lumigraph \cite{buehler2001unstructured} and generated novel views by interpolating from existing views. This approach required densely captured images for realistic rendering. Subsequently, some geometry-based methods were proposed. Mesh-based methods \cite{debevec2023modeling, riegler2020free, thies2019deferred, waechter2014let, aldinger2000surface} supported efficient rendering but struggled with surface optimization. Volume-based methods used voxel grids \cite{kutulakos2000theory, penner2017soft, seitz1999photorealistic} or multi-plane images (MPI) \cite{anantrasirichai2022artificial, mildenhall2019local, srinivasan2019pushing, zhou2018stereo}, offering detailed rendering but suffering from low memory efficiency or being limited to small view variations. In recent years, the method proposed by Neural Radiance Field (NeRF) \cite{mildenhall2021nerf} to model 3D scenes using MLP has gained widespread recognition. This has led to a series of related research efforts, such as speeding up training and inference speeds \cite{chen2022tensorf, fridovich2022plenoxels, muller2022instant}, improving rendering quality \cite{barron2021mip, verbin2022ref}, view synthesis for dynamic scenes \cite{pumarola2021d, cao2023hexplane, fridovich2023k}. However, these methods have consistently struggled to achieve real-time rendering. Last year, 3DGS  \cite{kerbl20233d} emerged and addressed this issue for people. Not only that, but 3DGS also offers higher rendering quality and interpretability. More and more people are getting involved in related research \cite{yu2023mip,huang20242d,lu2023scaffold}.

\section{Method}
\label{others}

Given a series of low dynamic range (LDR) images captured under multiple exposure conditions from different viewpoints, our task is to efficiently reconstruct a high-quality high dynamic range (HDR) radiance field solely from these LDR images and obtain HDR images through rendering. The entire framework is illustrated in the Fig \ref{fig:HDR-GS pipeline}. In this section, we will provide a detailed introduction to each component of our method. In Sec.\ref{M1}, we introduce the basic process of rendering and tone mapping. In Sec.\ref{M2}, we discuss the design of our tone mapper function. Then, in Sec.\ref{M3}, we introduce our coarse-to-fine strategy, followed by the optimization process in Sec.\ref{M4}.

\subsection{Preliminary}\label{sec:preliminary}
3DGS initializes with a sparse point cloud generated from Structure-from-Motion (SfM) \cite{buehler2001unstructured}. This method models geometric shapes as a set of 3D Gaussian functions defined by covariance matrices and means in world space.  
\begin{equation}
 G(x) = e^{-\frac{1}{2}(x-\mu_{3D})^T\Sigma_{3D}^{-1}(x-\mu_{3D})} 
\end{equation}
Where \( \mu_{3D} \) is the mean of the Gaussian point. To ensure the 3D Gaussian covariance matrix retains physical meaning, specifically to maintain positive semi-definiteness, it can be further decomposed into a scale matrix \( S \) and a rotation matrix \( R \). Thus, the 3D covariance matrix can be represented as:
\begin{equation}
    \Sigma_{3D}=RSS^{T}R^{T}
\end{equation}
where \( \Sigma_{3D} \) is the covariance matrix of the Gaussian point. To render the image, first approximate the projection of the 3D Gaussian into 2D image space using perspective transformation. Specifically, the projection of the 3D Gaussian is approximated as a 2D Gaussian with center \( \mu_{2D} \) and covariance \( \Sigma_{2D} \). \( \mu_{2D} \) and \( \Sigma_{2D} \) are computed as:
\begin{equation}
\mu_{2D} = (K ((W \mu_{3D})/(W \mu_{3D})_z))_{1:2}  ~~~~~ \Sigma_{2D} = (JW\Sigma_{3D} W^T J^T)_{1:2,1:2}   
\end{equation}
\(W\) and \(K\) are the  viewing transformation and projection matrix. Finally, after sorting the Gaussian points by depth, the pixel \(p\) value can be computed as:
\begin{equation}
\label{eq:3dgsrender}
    C_p=\sum_{i\in\mathcal{N}}\mathbf{c}_i\alpha_i\prod_{j=1}^{i-1}(1-\alpha_j), ~~~
    \text{where}~~~
    \alpha=\sigma\exp\left(-\frac{1}{2}(\mathbf{x}-\mu_{2D})^T\Sigma_{2D}^{-1}(\mathbf{x}-\mu_{2D})\right).
\end{equation}
\(c_i\) refers to the RGB color of Gaussian point which is represented by spherical harmonics (SH).

\subsection{Basic process}\label{M1}

\textbf{Radiance-based \(\alpha\) composition.} According to the principles of physical imaging, the radiance \(L\) emitted by objects in the scene is transformed into irradiance \(E\) on the surface of the image sensor as it passes through the camera lens. To simulate this process, we redefine the color of Gaussian points as radiance \(L\), which results in the pixel values of the image formed by Gaussian points splatting onto the HDR plane no longer representing color \(C\), but irradiance \(E\). We can rewrite the rendering formula of 3DGS\cite{kerbl20233d} \ref{eq:3dgsrender} as follows:
\begin{equation}
    E(\mathbf{p}) = \sum_{i =0 }^{N} L_i \alpha_i \prod_{j=1}^{i-1}(1-\alpha_j)~~ where~~ E(\mathbf{p})\in (0,+\infty)
\end{equation}
In which \(\mathbf{p}\) means the pixel, \(E\) represents pixel irradiance of \(\mathbf{p}\). as shown in Fig \ref{fig:HDR-GS pipeline}\(\textbf{(b)}\), \(L_i\) represents the radiance of i-th Gaussian point used to render pixel \(\mathbf{p}\), as shown in Fig \ref{fig:HDR-GS pipeline}\(\textbf{(a)}\).

\textbf{Imaging process.} The total irradiance received by the image sensor within the exposure time \(t\) results in the accumulated exposure. After undergoing photoelectric conversion by the sensor, analog-to-digital conversion, \textit{etc}. the LDR pixel value \(C\) is obtained. The entire imaging process can be represented by a function called the camera response function (CRF) \(F(\cdot)\). Combining \(E(\cdot)\), we represent the entire imaging process with the following formulation\cite{szeliski2022computer}:
\begin{equation}\label{eq: C =F(E t)}
    C( \mathbf{p},t) = F(E(\mathbf{p})*t(\mathbf{p}))
\end{equation}

where \( t \) represents the exposure time of the camera capturing a light ray (or pixel), which depends on the shutter speed. Following the classic non-parametric CRF calibration method by Debevec and Malik\cite{debevec2023recovering}, we assume that the CRF \(F(\cdot)\) is monotonic and invertible. Therefore, we can rewrite the Eq. \ref{eq: C =F(E t)} as:
\begin{equation}
    \ln F^{-1}(C(\mathbf{p},t)) = \ln E(\mathbf{p}) + \ln t(\mathbf{p})
\end{equation}

After further simplification, it can be expressed as the following equation:
\begin{equation}
\label{eq:4}
C(\mathbf{p},t)=g(\ln E(\mathbf{p})+\ln t(\mathbf{p})),\\where ~ g=(\ln F^{-1})^{-1}
\end{equation}

The tone mapper function can thus be transformed into the function \(g(\cdot)\),  where we use an asymmetric grid to model. We will illustrate the details of the asymmetric grid in Sec.\ref{M2}. Here, for convenience, we disregard the logarithm and denote \(\ln E(\mathbf{p}) + \ln t(\mathbf{p})\) as exposure. 

In practical operations, to streamline parameter learning and minimize the number of iterations required, we directly learn the value of \(\ln E\), the logarithmic domain of \(E\). Please note that the radiance of the Gaussian points learned after such transformation also change. We denote it as \( L'\).

\subsection{Grid-based tone mapper}\label{M2}

\begin{figure}
    \centering 
    \subfloat[]{\includegraphics[width=0.45\textwidth, trim=7cm 10.3cm 8cm 0cm, clip]{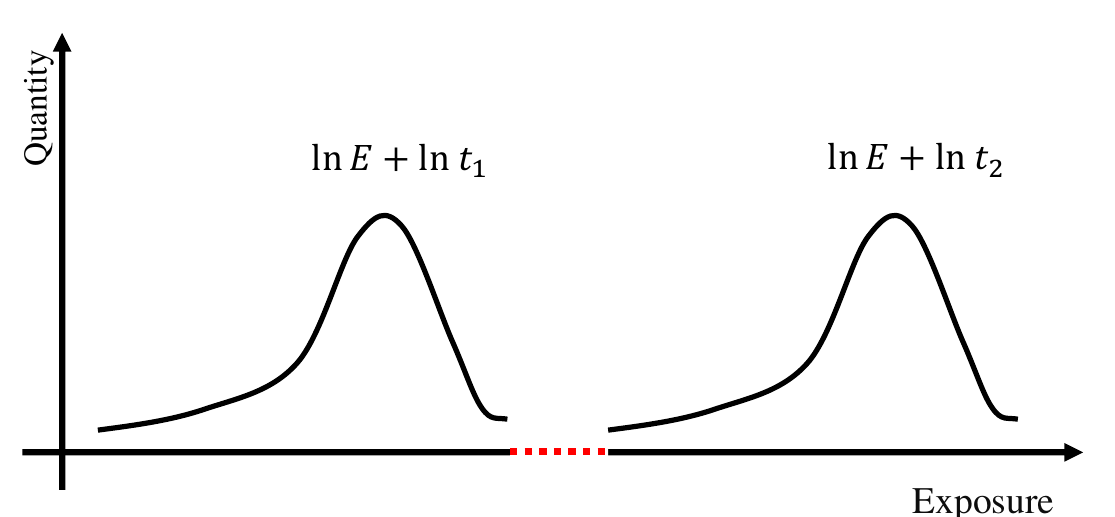}}
    \subfloat[]{\includegraphics[width=0.45\textwidth, trim=7cm 2.1cm 8cm 8.8cm, clip]{figs/t_2.pdf}}
    \caption{ \textbf{The distribution of accumulated exposure.} The x-axis represents the exposure values from a given camera viewpoint, and the y-axis represents the number of pixels with the corresponding exposure values. (a) represents the original distribution of accumulated exposure, while (b) represents the distribution after applying time scaling \(f_t (\cdot)\).} 
    \label{fig:t}
\end{figure}


To faithfully simulate the physical imaging process, in this section, we introduce our asymmetrical grid-based tone mapper \( g \)  to model the process from pixel irradiance \(E'\) to color \(c\), as shown in fig.\ref{fig:HDR-GS pipeline}(b). Recently, there have been several grid-based methods for modeling 3D scenes, such as \cite{yu2021plenoctrees, cao2023hexplane, fridovich2023k, fridovich2022plenoxels, muller2022instant}, \textit{etc}. These grid-based methods accelerate model training while ensuring rendering quality. Influenced by their work, we designed an asymmetric grid to model our tone mapper. The difference is that we use an asymmetric grid, while they use a symmetric one. We only need one tone mapping module to map irradiance to color for any seen or unseen viewpoint.

\textbf{Learned asymmetric grid.} We have empirically found that in some scenarios, the distribution of irradiance values is highly uneven. For instance, the majority of irradiance values might fall within the range (a, b), but a small portion may also lie within the range (b, c). If a uniform symmetric grid is used, modeling these two regions presents the following drawbacks: using a sparse grid in the range with dense irradiance distribution can lead to inaccurate mapping, while using a dense grid in the range with sparse irradiance distribution can cause overfitting. Therefore, as shown in Fig.\ref{fig:HDR-GS pipeline}(c), we introduce an asymmetric grid, which refers to a grid that lacks both central and density symmetry. In the range of values with a dense irradiance distribution (a,b), we use a dense grid (128 nodes per unit) to model the tonemapping function. Conversely, we use a sparse grid (64 nodes per unit) for regions with a sparse irradiance distribution (b,c). It offers greater flexibility and expressiveness compared to the symmetric grid \cite{yu2021plenoctrees, cao2023hexplane, fridovich2023k, fridovich2022plenoxels, jun2022hdr, muller2022instant}, especially in the scene with uneven irradiance distribution. Considering generality, we usually set b=0, but in some extreme cases, b needs to be adjusted according to the scenario.

In the training phase, gradients would not propagate backward if boundary values were merely assigned to the small percentage of exposure values that fall beyond the grid range.
Therefore, we design the following function to deal with these exposure values:
\begin{equation}\label{eq:g_leaky}
    g_{leaky}(x)=\left\{
    \begin{matrix}
        \beta(x-a), &x<a \\
        g(x), & a\leq x\leq b \\
        -\frac{\beta}{\sqrt{(x-b+1)}}+\beta+1, & b<x
    \end{matrix}
    \right.
\end{equation}
In this function, we usually set \(\beta = 0.01\). According to Debevec and Malik\cite{debevec2023recovering}, the asymmetric grid \(g\) also needs to satisfy \(g(a)=0\) and \(g(b)=1\). Rewriting the above Eq.\ref{eq:4}, we have:
\begin{equation}\label{eq:c3}
    C(\mathbf{p},t) = g_{leaky}(E_{ln}(\mathbf{p}) + t_{ln}(\mathbf{p}))
\end{equation}
Where \(E_{ln}\) and \(t_{ln}\) represent the logarithmic domains of \(E\) and \(t\) respectively.

\textbf{Exposure time scaling.} During experiments, directly optimizing the asymmetric grid \(g\) as described in Eq. \ref{eq:c3} presents considerable challenges. If the exposure time  of training dataset is \(\{t_1, t_2 \}\).  Given a camera viewpoint, the exposure distribution received by its camera plane may resemble that in Fig.\ref{fig:t}(a). There is a significant gap (marked in red) between the two curves is evident. This results in unoptimized regions in the grid \( g \) during the training phase. Therefore, when inferring LDR images with an exposure time of \( t_{1.5} \) (where \( t_1 < t_{1.5} < t_2 \)), errors occur because the grid region (marked in red) for exposure \( t_{1.5} \)  has not been trained. Consequently, we apply scaling and an offset to the logarithm of the exposure time \(t\) to mitigate these issues:
\begin{equation}
    t' = f_t (t) =r \cdot t_{ln} + s
\end{equation}
We determine \( r \) based on the exposure time, and if the exposure levels in the training data are too high, we need to scale them. Given a training set with monotonically increasing exposure times \(\{t_1, t_2, t_3, \ldots, t_n\}\), the formulas for determining \( r \) and \( s \) is:
\begin{equation}
 r = \min_{i} \left\{ (\frac{2t_{i}}{t_{i+1}})\right\}  ~~~~ s = -\frac{r\times(\ln t_{\text{max}} + \ln t_{\text{min}})}{2}.
\end{equation}
where $ t_{\text{max}} $ and $ t_{\text{min}} $ are the maximum and minimum values of exposure time $ t_i $. After the above processing, the distribution of exposure changes from Fig.\ref{fig:t}(a) to Fig.\ref{fig:t}(b). In Fig.\ref{fig:t}(b), the variance of exposure time is reduced, and the two independent areas converge and merge into one, resolving the issues of separate optimization and eliminating unoptimized zone.


Another noteworthy point is that applying the scaling and offset to \(t_{ln}\) will result in a change in the originally expected \(E_{ln}\). Instead of learning \(E_{ln}\), the model learns another pixel irradiance value, denoted as \(E'\). The relationship between \(E'\) and  \(E\) will be explained shortly. Rewriting the Eq.\ref{eq:c3}, the relationship between pixel irradiance \(E\) and LDR color \(C\) can be expressed as:
\begin{equation}\label{eq:c4}
    C(\mathbf{p},t) = g_{leaky}(E'(\mathbf{p}) + t'(\mathbf {p}))
\end{equation}

This equation is straightforward yet powerful, capturing both our implementation approach and the underlying physical imaging process. The success of our method is largely attributed to the ability to mathematically describe each module's design. As illustrated in Fig.\ref{fig:HDR-GS pipeline}(d), this equation enables our model to render LDR images across a range of exposure times.


\textbf{HDR mapping function.} As illustrated in Fig.\ref{fig:HDR-GS pipeline}(b), the model iteratively updates and learns \(E'\) via gradient backpropagation. However, \( E'\)  is not the desired HDR value \(E\). They are related through a mapping function  defined as follows:
\begin{equation}
    E =f_H(E') = e ^ {( (E'+ s)/r )}
\end{equation}
Please refer to the Supplement for more proof details. Through training, we obtain the corresponding HDR image \(I_h\), where each pixel value represents HDR value \(E\).

\subsection{Coarse to fine optimization strategy}\label{M3}

The above sections describe the main process of our training pipe. However, we empirically found that directly using a grid as the tonemapping function and jointly training it with the attributes of Gaussian points leads to severe coupling and overfitting to the training dataset, as shown in Fig.\ref{fig:ablation}. Therefore, we designed a coarse-to-fine strategy: during the coarse phase, we use a fixed function \(g_s\) as the tonemapper and train only the attributes of the Gaussian points. In the fine phase, we then use the grid as the tonemapper and jointly train it with the attributes of the Gaussian points. Now, the next question is which function \(g_s\) we should choose as the tonemapper during the coarse phase. According to Debevec \cite{debevec2023recovering}, we can opt for a monotonically increasing, smooth function bounded between 0 and 1 as our tone mapper for pre-training the Gaussian points. For simplicity, we opt for the sigmoid function. Therefore, in the coarse phase, the function relationship between pixel irradiance \(E'\) and LDR color \(C\) can be expressed as:
\begin{equation}
    C(\mathbf{p},t) = g_{s}(E'(\mathbf{p}) + t'(\mathbf {p}))
\end{equation}
During the coarse phase, we typically set the number of training iterations to 6000, and the training takes approximately 50 seconds. We also noticed that \cite{wu2024fast} solely employs a sigmoid function as their tone mapper. Due to the limited expressive power of the sigmoid function, this can lead to their HDR images experiencing floating-point issues and biasing towards white. Lastly, we will highlight the significance of the coarse stage in ablation research Tab.\ref{tab:Ablation study}.

\subsection{Loss function}\label{M4}

\textbf{Reconstruction Loss.} We adopt the same loss function as in 3DGS \cite{kerbl20233d} to constrain our Gaussian points:
\begin{equation}
     \mathcal{L}_{rec} = (1-\lambda_1)\mathcal{L}_{1} + \lambda_1\mathcal{L}_{D-SSIM}, 
\end{equation}

\textbf{Smooth Loss:} To ensure that our grid \( g \) conforms to the CRF properties proposed by Debevec \cite{debevec2023recovering} that CRFs increase smoothly, we employ the loss function \cite{jun2022hdr}:
\begin{equation}
    \mathcal{L}_{smooth} = \sum_{i=1}^N \sum_{e \in [a,b]} g_i''(e)^2,
\end{equation}
Where the \(g_i''(e)\) denotes the second order derivative of grid \textit{w.r.t.} \(e\) in the domain \((a,b)\) of the grid.   

\textbf{Unit Exposure Loss.} Learning HDR radiance fields solely from LDR images may result in radiance fields with various scales \( \alpha E \). To ensure consistency between the HDR radiance fields we construct and those generated by Blender, facilitating HDR quality evaluation, we employ the loss function introduced by HDRNeRF\cite{huang2022hdr}:
\begin{equation}
     \mathcal{L}_u = \| g(0) - C_0 \|_2^2.
\end{equation}

Where \(C_0 = 0.73\). The ablation studies shown in Table. \ref{tab:Ablation study} and Fig. \ref{fig:compair ldr}. 

\textbf{Total loss.} Finally, by combining the aforementioned loss functions, we derive the total loss function as follows:
\begin{equation}
    \mathcal{L}_{total} = \mathcal{L}_{rec} + \lambda_{2} \mathcal{L}_{smooth} + \lambda_3\mathcal{L}_u 
\end{equation}

\section{Experiments}
In this section, we will present our experimental setup and the dataset used in Sec.\ref{imple details} and \ref{dataset} respectively. We will go into more detail about the experiment results in Sec.\ref{Evaluaciton}. Our experiments on synthetic and real datasets demonstrate that our method achieves state-of-the-art performance.

\subsection{Implementation details} 
\label{imple details}
To avoid the explosion of point numbers during training in complex scenes, we employ the pruning strategy proposed by \cite{niemeyer2024radsplat}, as detailed in the supplementary material. The Gaussian point parameters are set in the same way as in 3DGS\cite{kerbl20233d}.  The loss function parameters for synthetic datasets are \( \lambda_{2} = 0.3, \lambda_3 = 0.5 \), but for real datasets they are \( \lambda_{1} = 0.2, \lambda_{2} = 1e-3, \lambda_3 = 0 \). We utilize the Adam optimizer\cite{kingma2014adam} for training. As for the learning rate of the asymmetric grid, we initially set it to 0.02, and then decayed to 5e-6. Our usual setting in sparsely interpolated areas is less than 64 nodes. The entire model is trained on an NVIDIA A100, with 6000-14000 epochs for the coarse phase, completing in under 1 minute, and 17000-30000 epochs for the fine phase, totaling approximately 4-8 minutes of training time. We also test our code on the Tesla V100, and the training time does not exceed 10 minutes. Our model runs consistently with a GPU memory use of less than 5GB during training.

\subsection{Dataset}
\label{dataset}
The dataset we utilize is provided by HDRNeRF\cite{huang2022hdr}, consisting of 8 synthetic scenes rendered by Blender and 4 real scenes captured by digital cameras. Each dataset comprises 35 different viewpoints captured by a forward-facing camera, with each viewpoint having 5 exposure levels \(\{t_1,t_2,t_3,t_4,t_5\}\), ranging from -4EV to 5EV. We strictly followed HDRNeRF's training and measurement protocols, employing one image per exposure level for each view, resulting in a total of 18 training images. The test images consist of 85 images, covering 5 exposure levels across 17 views. The resolution of each view in synthetic scenes is 400 × 400 pixels, whereas in real scenes, it is 804 × 534 pixels.



\begin{table}[t]
    \centering
    \caption{Quality Comparison. LDR-OE represents the use of LDR images with exposures \(\{t_1,t_3,t_5\}\) as the training set, while LDR-NE represents the use of LDR images with exposures \(\{t_2,t_4\}\) as the training set. The exposure times of the test dataset are \( \{ t_1, t_2, t_3, t_4, t_5 \} \). HDR denotes the HDR results. The \textcolor{red}{best} and the \textcolor{blue}{second best} results are denoted by red and blue.}
    \label{all quality comparison}
    \begin{threeparttable}
    \resizebox{\textwidth}{!}
    {
        \begin{tabular}{c|c|ccccccccccc}
        \toprule[2pt]
        \multicolumn{2}{c|}{} & \multicolumn{3}{c}{LDR-OE} & \multicolumn{3}{c}{LDR-NE} & \multicolumn{3}{c}{HDR} & \multicolumn{2}{c}{} \\
        \multicolumn{2}{c|}{} & PSNR \(\uparrow\)  & SSIM \(\uparrow\)  & LPIPS \(\downarrow\)   & PSNR\(\uparrow\)  & SSIM\(\uparrow\)  & LPIPS \(\downarrow\) & HDR VDP-Q\(\uparrow\)  & PUPSNR\(\uparrow\)  & PUSSIM \(\uparrow\)  & FPS\(\uparrow\)  & Time \(\downarrow\) \\
        
        \midrule[1pt]
        \multirow{2}{*}{NeRF } 
        & syn.  & 13.97 & 0.555 & 0.376 & — & — & —&— &— &— \\
        & real. &  14.95 & 0.661& 0.308&  — & — & —&— &— &— \\ 
        \midrule[0.5pt]
        \multirow{2}{*}{NeRF-W } 
        & syn. & 29.83 & 0.936& 0.047 &29.22& 0.927 &\textcolor{blue}{0.050} &— &— &— \\
        & real. &28.55 &0.927 &0.094 &28.64& 0.923 &0.089& — &—& — \\ 


        \midrule[0.5pt]
        \multirow{2}{*}{HDR-NeRF} 
        & syn. & \textcolor{blue}{36.72} &\textcolor{blue}{0.951} & \textcolor{blue}{0.044} &\textcolor{blue}{34.43} &\textcolor{blue}{0.948} & {0.064} & \textcolor{blue}{6.56} &\textcolor{blue}{20.33}& \textcolor{blue}{0.527} & <1 & 9.4 hours \\
        
        & real. &\textcolor{blue}{31.88} & \textcolor{blue}{0.950} &\textcolor{blue}{0.067} &\textcolor{blue}{30.43} &\textcolor{blue}{0.946} &\textcolor{blue}{0.078} & — &—& — & <1 & 8.2 hours \\
        \midrule[0.5pt]
        \multirow{2}{*}{3DGS}
        & syn. &11.24 &0.395 &	0.534 &  11.97 &0.431 &	0.489   &— &— &—& \textcolor{red}{352.5}& 	\textcolor{blue}{12.7 mins }
\\
        & real. &11.87 &0.611 	&0.355 & 12.99 	&0.682 &0.339 &—& —& — &\textcolor{red}{391.7} &\textcolor{blue}	{11.3 mins }
 \\
        
        \midrule[0.5pt]
        \multirow{2}{*}{Ours} 
        & syn. & \textcolor{red}{39.16} &	\textcolor{red}{0.974} &\textcolor{red}{0.012}&\textcolor{red} {38.84} &	\textcolor{red}{0.975}&\textcolor{red}{0.012} &\textcolor{red} {9.70} &\textcolor{red}{23.67}& \textcolor{red}{0.835} & \textcolor{blue}{248.8 }& \textcolor{red}{6.8 mins}  \\
        
        & real. & \textcolor{red}{33.34}  &	\textcolor{red}{0.967}  & \textcolor{red}{0.023}  & \textcolor{red}{32.77}  &\textcolor{red}{0.960}  &\textcolor{red}{0.030}  & — &—& — & \textcolor{blue}{210.9} & \textcolor{red}{8.5 mins}
        
        \\
        
        \bottomrule[2pt]
        \end{tabular}
    }
\end{threeparttable}
\end{table}

\begin{figure}[t]
\centering
\rotatebox{90}{\scriptsize{~~~~~~~~~~~~~~bear}}\hspace{0.1cm}
\begin{minipage}[b]{0.16\columnwidth}
		\subfloat{\includegraphics[scale=0.16]{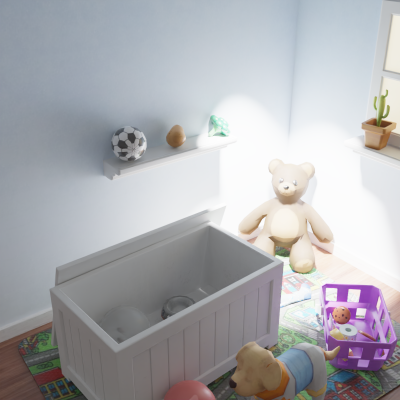}}
\end{minipage} 
\begin{minipage}[b]{0.16\columnwidth}
		\subfloat{\includegraphics[scale=0.16]{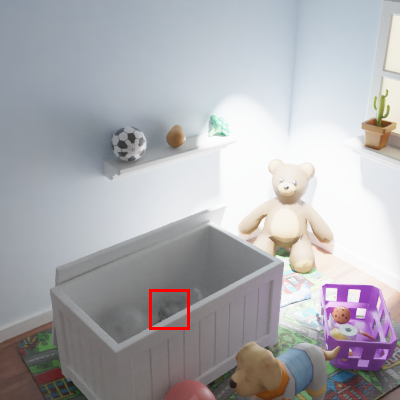}}
\end{minipage} 
\begin{minipage}[b]{0.08\columnwidth}
		\subfloat{\includegraphics[scale=0.158]{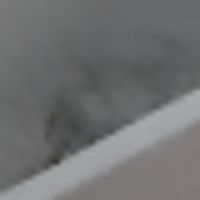}}
		\\
		\subfloat{\includegraphics[scale=0.158]{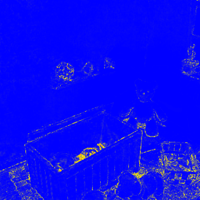}}
\end{minipage}
\begin{minipage}[b]{0.16\columnwidth}
		\subfloat{\includegraphics[scale=0.16]{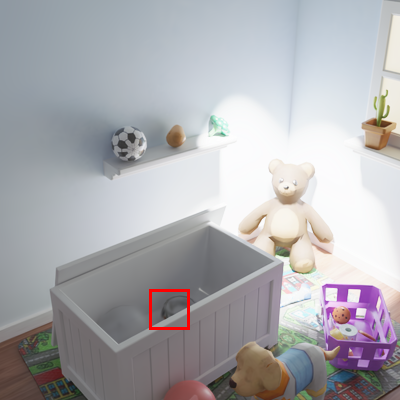}}
\end{minipage} 
\begin{minipage}[b]{0.08\columnwidth}
		\subfloat{\includegraphics[scale=0.158]{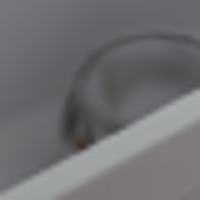}}
		\\
		\subfloat{\includegraphics[scale=0.158]{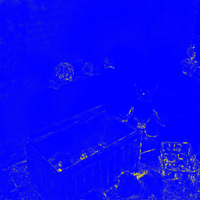}}
\end{minipage}
\begin{minipage}[b]{0.16\columnwidth}
		\subfloat{\includegraphics[scale=0.16]{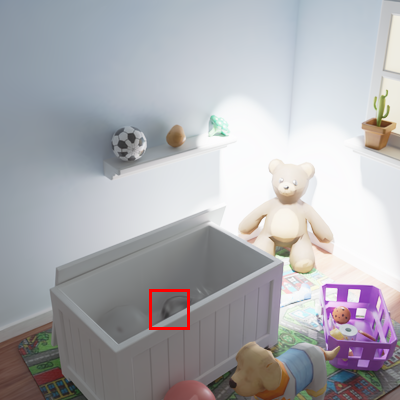}}
\end{minipage} 
\begin{minipage}[b]{0.08\columnwidth}
		\subfloat{\includegraphics[scale=0.158]{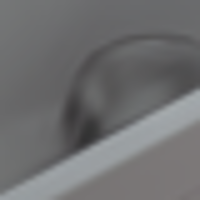}}
		\\
		\subfloat{\includegraphics[scale=0.158]{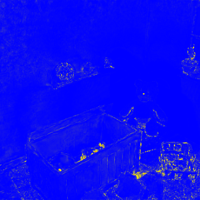}}
\end{minipage}

\rotatebox{90}{\scriptsize{~~~~~~~~~~~~~~desk}}\hspace{0.1cm}
\begin{minipage}[b]{0.16\columnwidth}
		\subfloat{\includegraphics[scale=0.16]{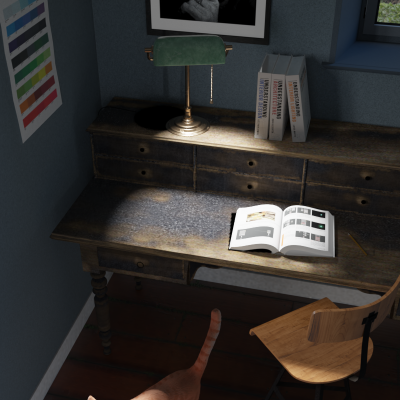}}
\end{minipage} 
\begin{minipage}[b]{0.16\columnwidth}
		\subfloat{\includegraphics[scale=0.16]{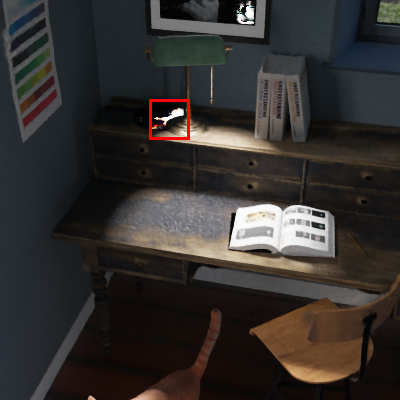}}
\end{minipage} 
\begin{minipage}[b]{0.08\columnwidth}
		\subfloat{\includegraphics[scale=0.158]{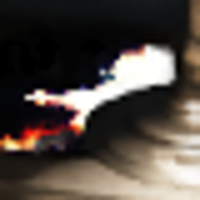}}
		\\
		\subfloat{\includegraphics[scale=0.158]{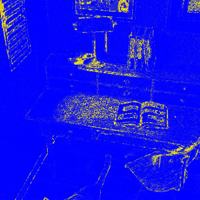}}
\end{minipage}
\begin{minipage}[b]{0.16\columnwidth}
		\subfloat{\includegraphics[scale=0.16]{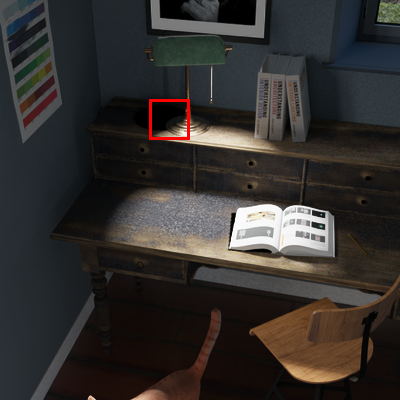}}
\end{minipage} 
\begin{minipage}[b]{0.08\columnwidth}
		\subfloat{\includegraphics[scale=0.158]{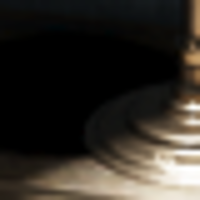}}
		\\
		\subfloat{\includegraphics[scale=0.158]{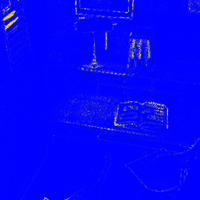}}
\end{minipage}
\begin{minipage}[b]{0.16\columnwidth}
		\subfloat{\includegraphics[scale=0.16]{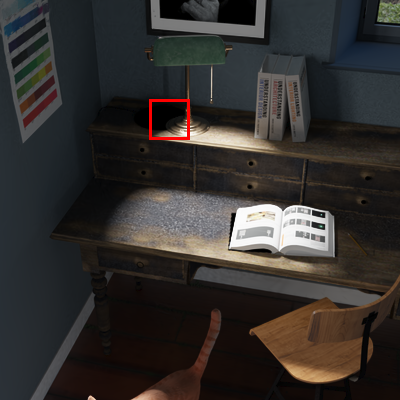}}
\end{minipage} 
\begin{minipage}[b]{0.08\columnwidth}
		\subfloat{\includegraphics[scale=0.158]{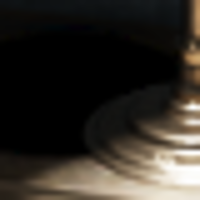}}
		\\
		\subfloat{\includegraphics[scale=0.158]{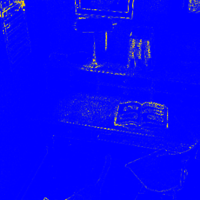}}
\end{minipage}

\rotatebox{90}{\scriptsize{~~~~~~~~~~~~~~dog}}\hspace{0.05cm}
\begin{minipage}[b]{0.16\columnwidth}
		\subfloat{\includegraphics[scale=0.16]{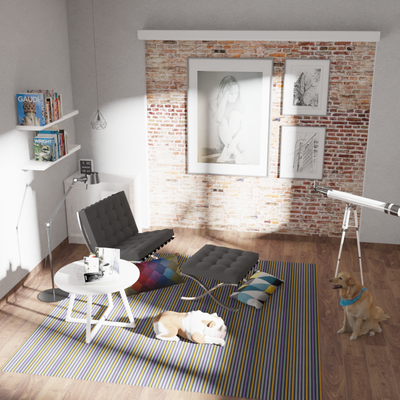}}
\end{minipage} 
\begin{minipage}[b]{0.16\columnwidth}
		\subfloat{\includegraphics[scale=0.16]{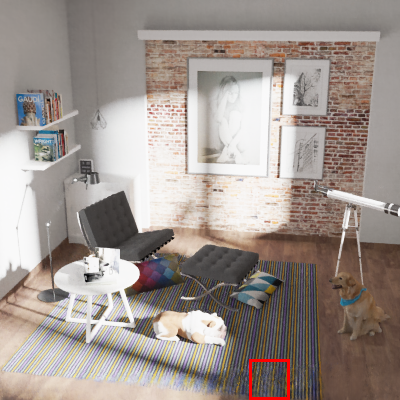}}
\end{minipage} 
\begin{minipage}[b]{0.08\columnwidth}
		\subfloat{\includegraphics[scale=0.158]{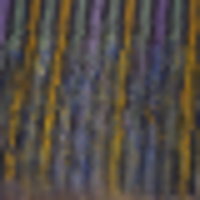}}
		\\
		\subfloat{\includegraphics[scale=0.158]{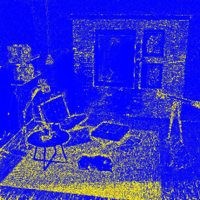}}
\end{minipage}
\begin{minipage}[b]{0.16\columnwidth}
		\subfloat{\includegraphics[scale=0.16]{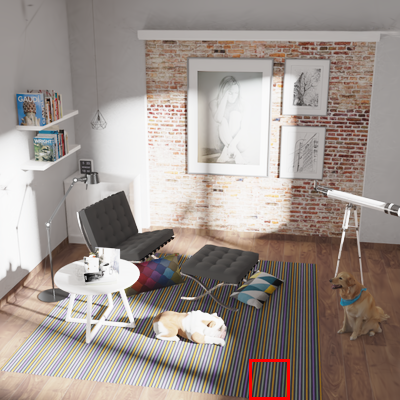}}
\end{minipage} 
\begin{minipage}[b]{0.08\columnwidth}
		\subfloat{\includegraphics[scale=0.158]{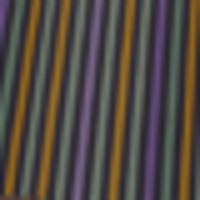}}
		\\
		\subfloat{\includegraphics[scale=0.158]{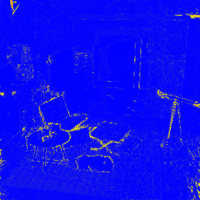}}
\end{minipage}
\begin{minipage}[b]{0.16\columnwidth}
		\subfloat{\includegraphics[scale=0.16]{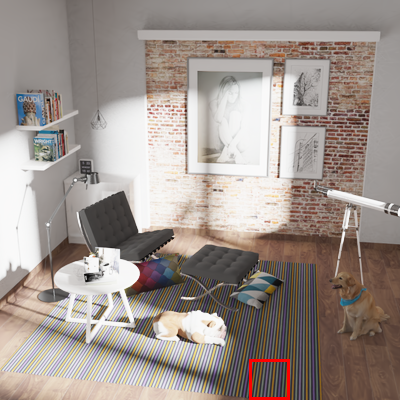}}
\end{minipage} 
\begin{minipage}[b]{0.08\columnwidth}
		\subfloat{\includegraphics[scale=0.158]{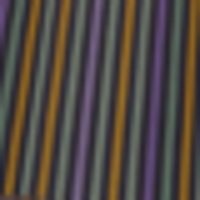}}
		\\
		\subfloat{\includegraphics[scale=0.158]{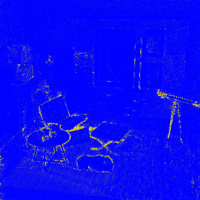}}
\end{minipage}

\setcounter{subfigure}{0}  

\rotatebox{90}{\scriptsize{~~~~~~~~~~~~~sponza}}\hspace{0.1cm}
\begin{minipage}[b]{0.16\columnwidth}
		\subfloat[GT]{\includegraphics[scale=0.16]{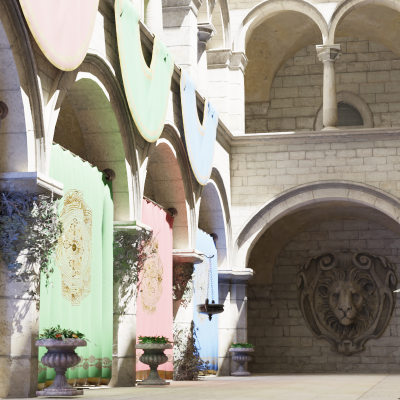}}
\end{minipage} 
\begin{minipage}[b]{0.16\columnwidth}
		\subfloat[HDR-NeRF]{\includegraphics[scale=0.16]{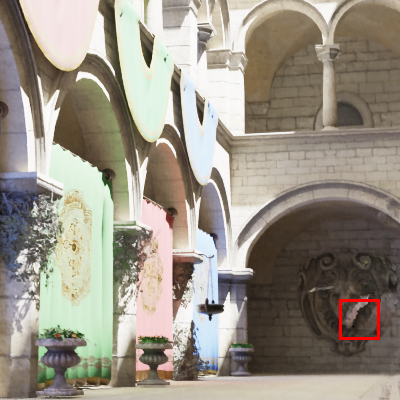}}
\end{minipage} 
\begin{minipage}[b]{0.08\columnwidth}
		\subfloat{\includegraphics[scale=0.158]{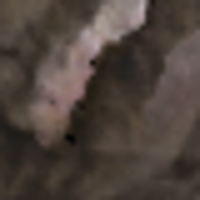}}
		\\
		\subfloat{\includegraphics[scale=0.158]{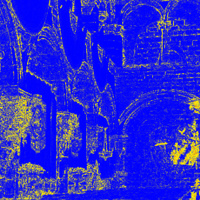}}
\end{minipage}
\begin{minipage}[b]{0.16\columnwidth}
\setcounter{subfigure}{2}  
\subfloat[Ours*]
  {\includegraphics[scale=0.16]{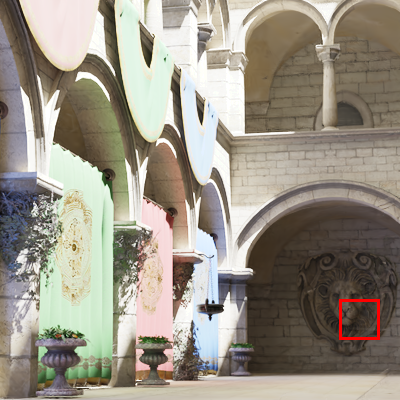}}
\end{minipage} 
\begin{minipage}[b]{0.08\columnwidth}
		\subfloat{\includegraphics[scale=0.158]{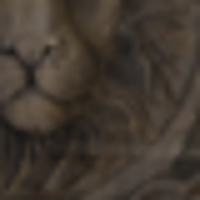}}
		\\
		\subfloat{\includegraphics[scale=0.158]{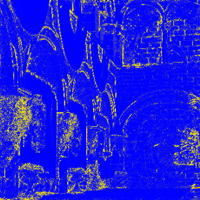}}
\end{minipage}
\begin{minipage}[b]{0.16\columnwidth}
\setcounter{subfigure}{3}  
		\subfloat[Ours]{\includegraphics[scale=0.16]{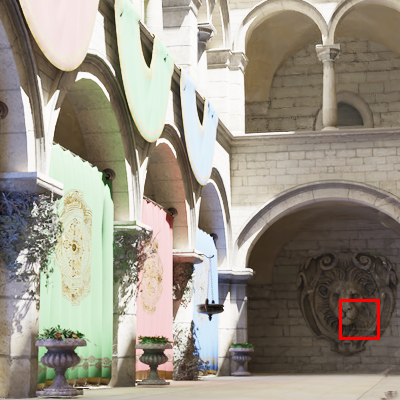}}
\end{minipage} 
\begin{minipage}[b]{0.08\columnwidth}
		\subfloat{\includegraphics[scale=0.158]{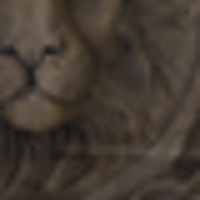}}
		\\
		\subfloat{\includegraphics[scale=0.158]{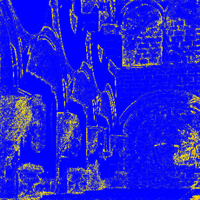}}
\end{minipage}
\caption{Comparison of LDR Image Quality for Novel View Renderings. The image in the top right of each rendered image is a zoomed-in section selected by the red rectangle. The mse error heatmap for each rendered image is shown in the bottom right. "Ours*" represents our method without \(\mathcal{L}_u\)}
\label{fig:compair ldr}
\end{figure}

\subsection{Evaluation}
\label{Evaluaciton}

We will compare our method with the following baseline methods: NeRF  \cite{mildenhall2021nerf}, NeRF-w \cite{martin2021nerf}, HDRNeRF \cite{huang2022hdr}, HDR-Plenoxel \cite{jun2022hdr}, and 3DGS \cite{kerbl20233d}. Additionally, we conducted ablation studies on some of our key modules, as presented in Table. \ref{tab:Ablation study}. The metrics measured include training time, PSNR, SSIM \cite{wang2003multiscale}, LPIPS \cite{zhang2018unreasonable}, High Dynamic Range-Visual Difference Predictor (HDR-VDP) \cite{mantiuk2004visible, mantiuk2011hdr, mantiuk2023hdr, narwaria2015hdr}, PUPSNR \cite{azimi2021pu21}, PUSSIM \cite{azimi2021pu21}, and FPS. All the HDR results are tone-mapped using Photomatix. These measurement results collectively indicate the efficiency of our proposed method.

\begin{figure}[t]
\centering
\subfloat{
		\includegraphics[scale=0.12]{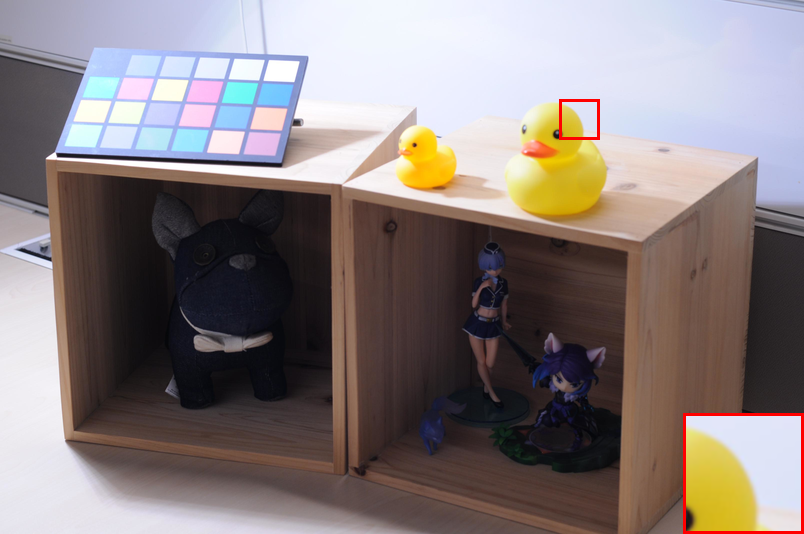}}
\subfloat{
		\includegraphics[scale=0.12]{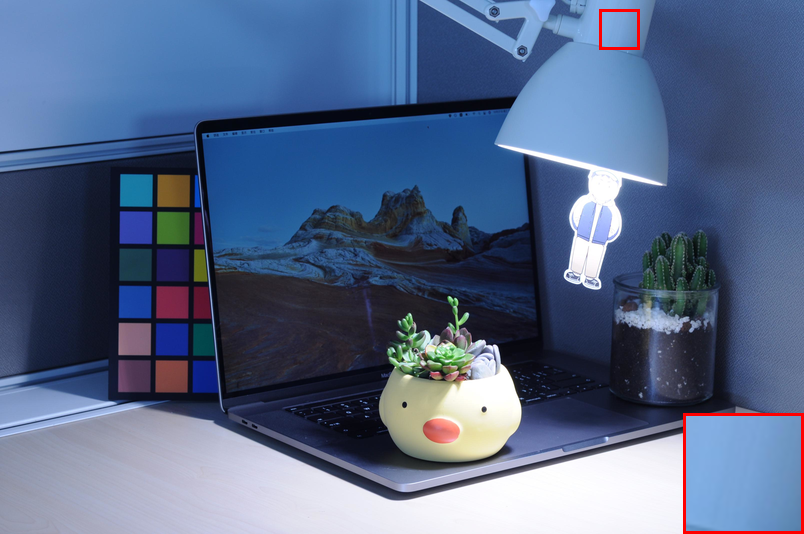}}
\subfloat{
		\includegraphics[scale=0.12]{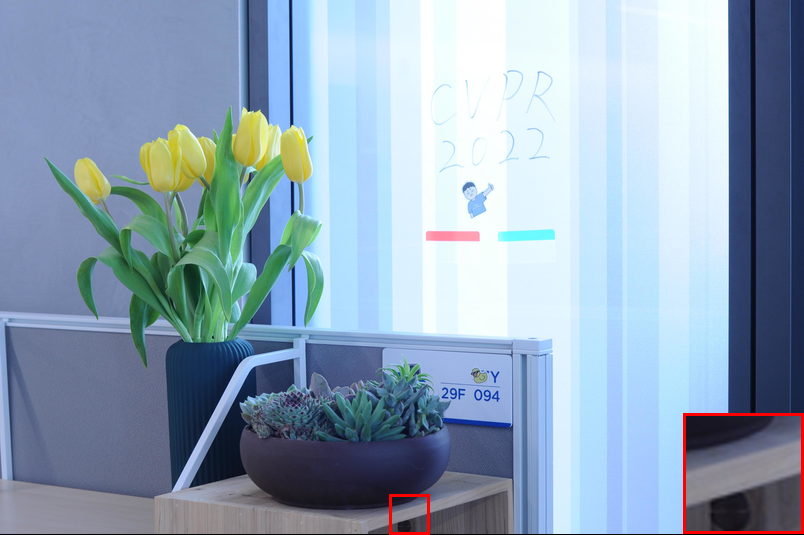}}
\subfloat{
		\includegraphics[scale=0.12]{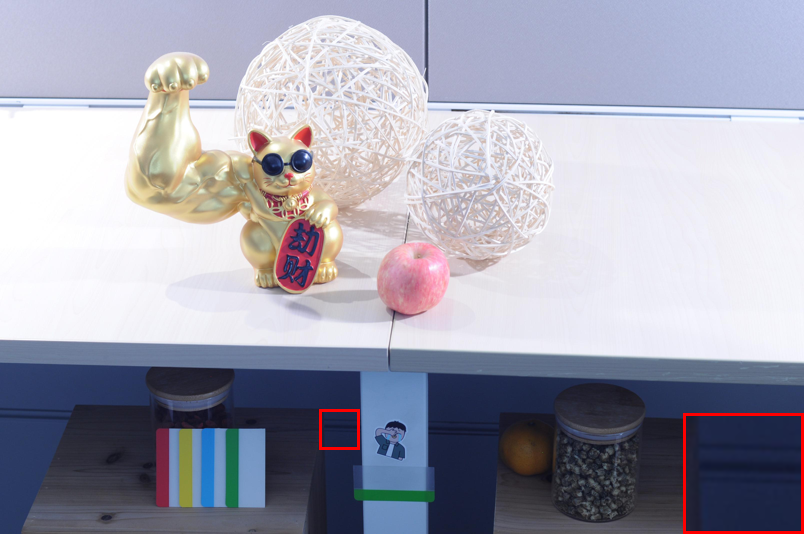}}

\subfloat{
		\includegraphics[scale=0.12]{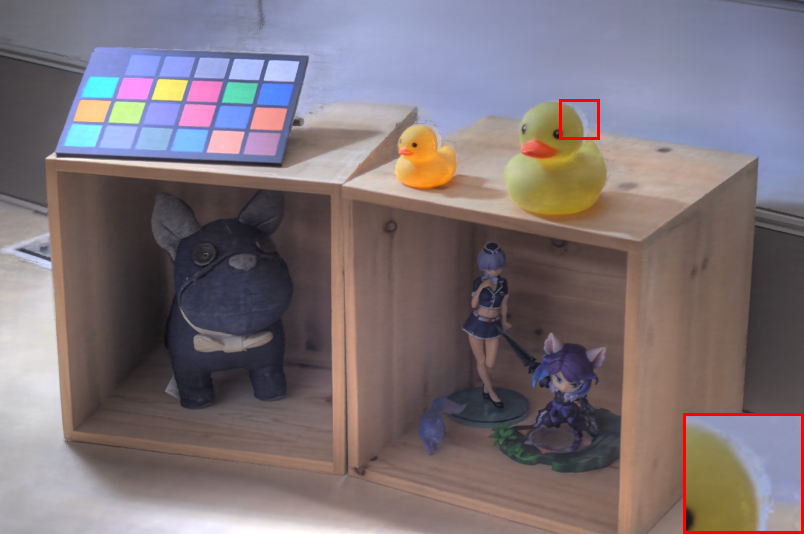}}
\subfloat{
		\includegraphics[scale=0.12]{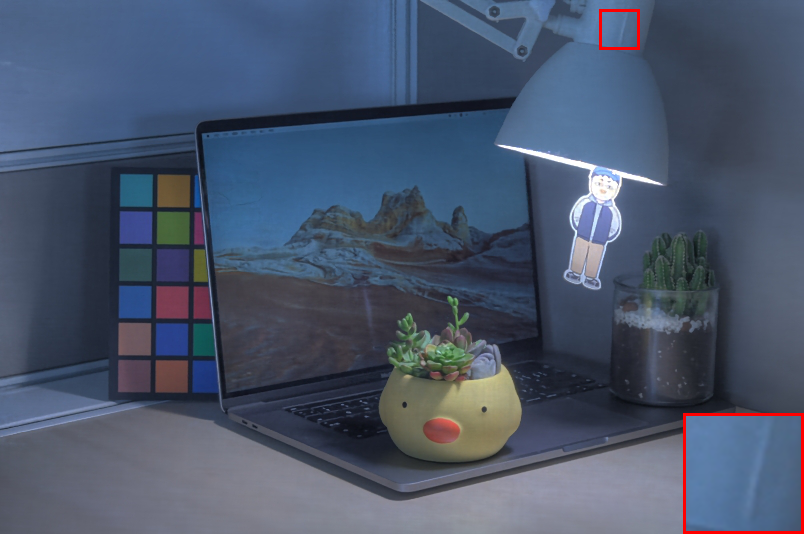}}
\subfloat{
		\includegraphics[scale=0.12]{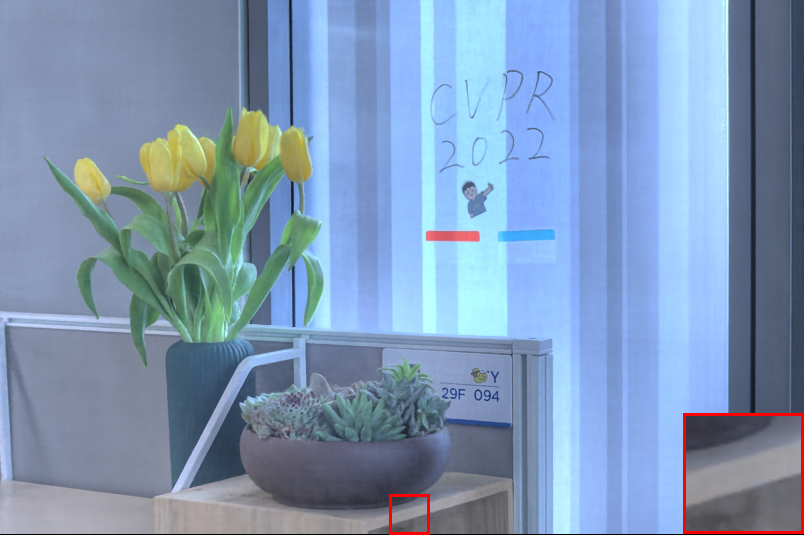}}
\subfloat{
		\includegraphics[scale=0.12]{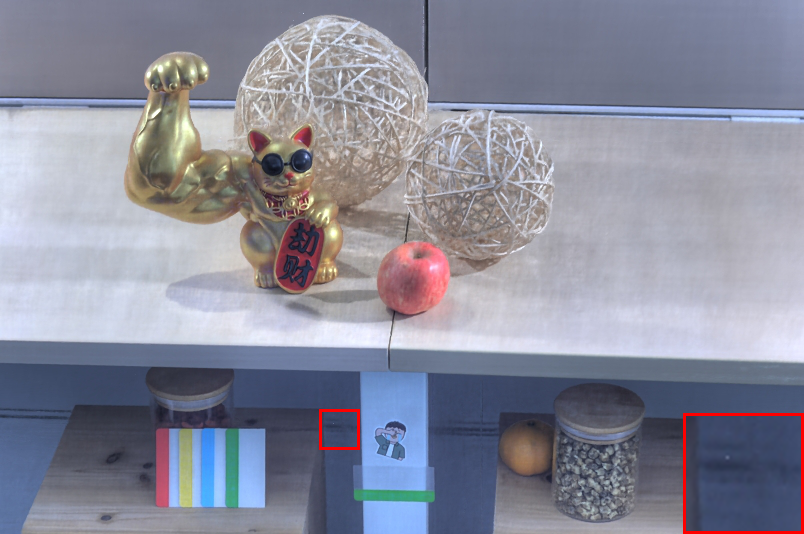}}

\setcounter{subfigure}{0}  
\subfloat[box]{
		\includegraphics[scale=0.12]{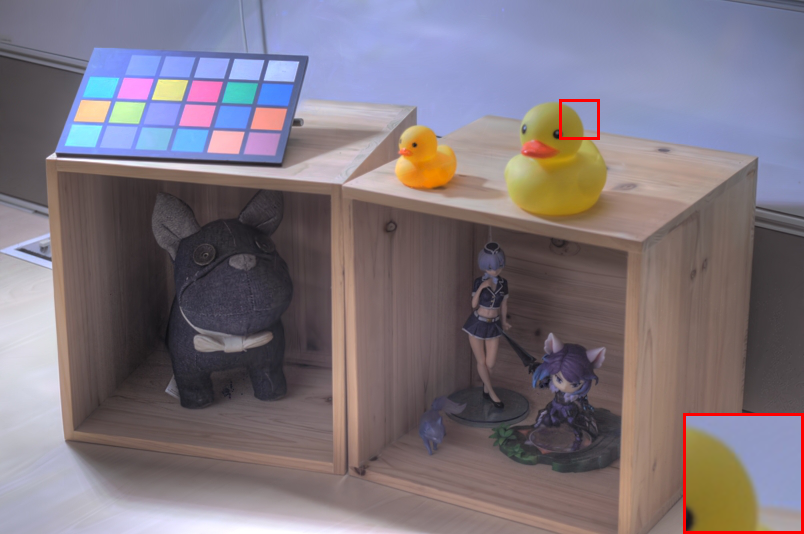}}
\subfloat[computer]{
		\includegraphics[scale=0.12]{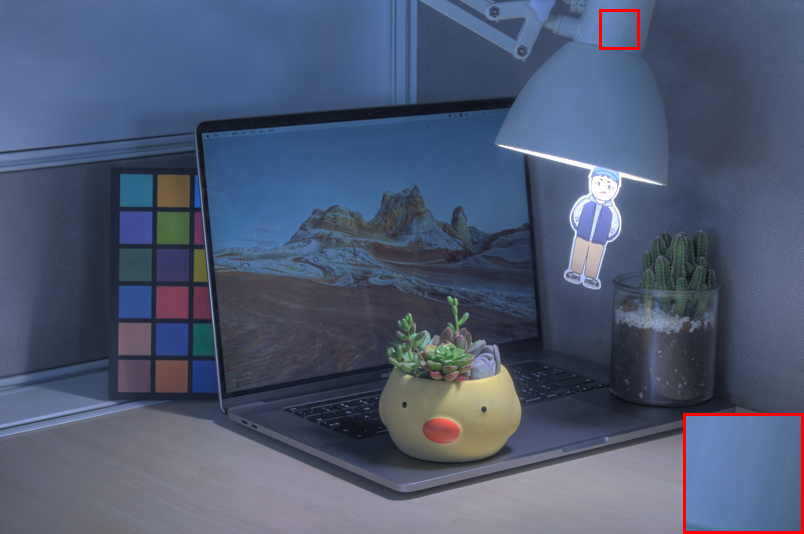}}
\subfloat[flower]{
		\includegraphics[scale=0.12]{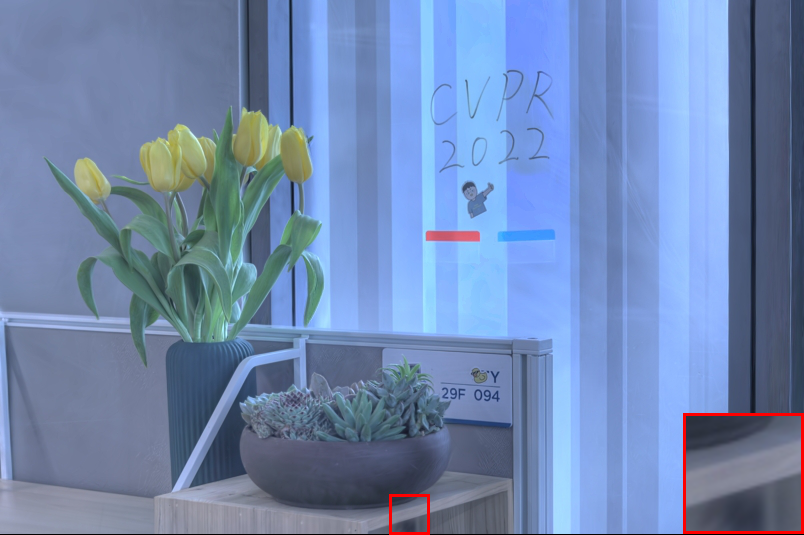}}
\subfloat[luckycat]{
		\includegraphics[scale=0.12]{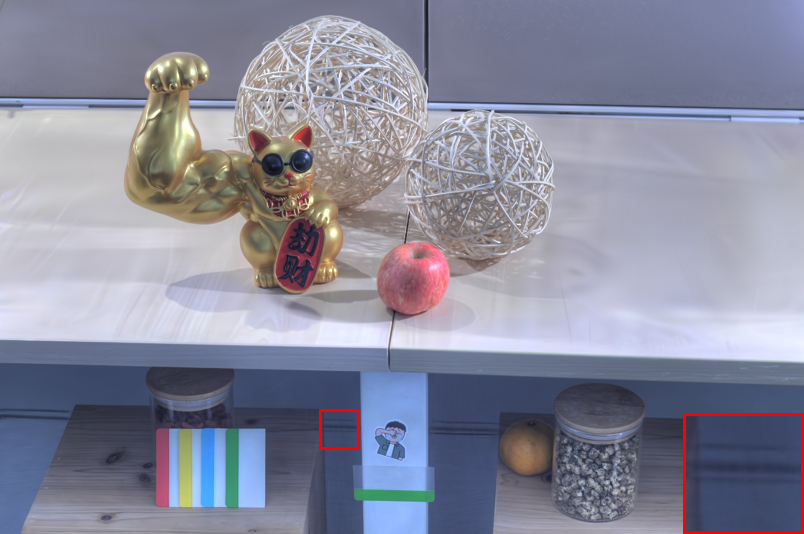}}

\caption{Comparison of HDR image quality for novel viewpoints on real scenes. The first row is GT LDR images, the second row HDR images are rendered by HDRNeRF\cite{huang2022hdr} , and the third row is ours.}
\label{fig:compair hdr real}
\end{figure}

\textbf{HDR metrics.} We employ two Authoritative HDR image measurement methods: High Dynamic Range-Visual Difference Predictor (HDR-VDP) \cite{mantiuk2004visible, mantiuk2011hdr, mantiuk2023hdr, narwaria2015hdr} and Perceptually Uniform (PU21) \cite{azimi2021pu21}. HDR-VDP simulates the human eye's perception of high dynamic range images under various lighting conditions, offering a means to quantify perceptual differences in images. We have opted for version 3.0.7 of the HDR-VDP measurement tool, employing the Side by Side assessment approach and the Q-score measurement metric. In this metric, a score of 10 indicates the highest quality, which diminishes for lower quality. PU21 \cite{azimi2021pu21} propose introducing a PU21 encoder to map the HDR pixel value range to the LDR pixel value range, thereby enabling the utilization of conventional LDR measurement techniques (such as PSNR, SSIM, LPIPS) for assessment. During PU21 testing, the maximum luminance of the HDR display is set to 1000 nits.

\begin{wraptable}{r}{7cm}
\centering
\vspace{-1.2em}
\caption{Some ablation studies of our method} 
\label{tab:Ablation study}
\resizebox{0.5\textwidth}{!}
{\begin{tabular}{cccc|cccccc}
\toprule[2pt]
 \multirow{1}{*}{\( \mathcal L_u \)}  & \multirow{1}{*}{Coarse} & \multirow{1}{*}{Time scaling} & \multirow{1}{*}{Asymmetric}  & \multicolumn{1}{c}{PSNR}\(\uparrow\) & \multicolumn{1}{c}{SSIM} \(\uparrow\) & \multicolumn{1}{c}{LPIPS} \(\downarrow\) \\
\midrule[1pt]
\CheckmarkBold  & \CheckmarkBold & \CheckmarkBold & \CheckmarkBold & 39.00 & 0.975 & 0.012  \\
\XSolidBrush  & \CheckmarkBold & \CheckmarkBold & \CheckmarkBold & 38.76  & 0.976 & 0.012  \\
\CheckmarkBold & \XSolidBrush & \CheckmarkBold & \CheckmarkBold & 19.32 & 0.618 & 0.308 \\
\CheckmarkBold & \CheckmarkBold & \XSolidBrush & \CheckmarkBold & 26.00 & 0.785 & 0.180   \\
\CheckmarkBold & \CheckmarkBold & \CheckmarkBold & \XSolidBrush & 38.60 & 0.975 & 0.013   \\
\bottomrule[2pt]
\end{tabular}}

\end{wraptable}


\textbf{Quality comparison.} The comparison results of our method with other methods on synthetic and real datasets are presented in Table \ref{all quality comparison}. It is evident that our method exhibits even faster training speeds compared to 3DGS \cite{kerbl20233d}. This is attributed to the tendency of the 3DGS method to overfit to the LDR images from each viewpoint, leading to an excessive number of points and consequently slowing down the training speed. Moreover, we adopt a pruning strategy \cite{niemeyer2024radsplat} to reduce the number of points, thereby enhancing the training speed. Furthermore, it can be observed that our method exhibits significantly better rendering quality compared to the second best—HDRNeRF, especially in LDR-NE (using only two exposure times as the training set). Unlike HDRNeRF, our method achieves high rendering quality in both LDR-NE and LDR-OE (using three exposure times as the training set), demonstrating greater stability. More importantly, our method can achieve real-time rendering, and the training time only takes several minutes.

\begin{figure}[t]
\centering
\rotatebox{90}{\scriptsize{~~~~~~~~~~~~~bathroom}}
\subfloat{
		\includegraphics[scale=0.185]{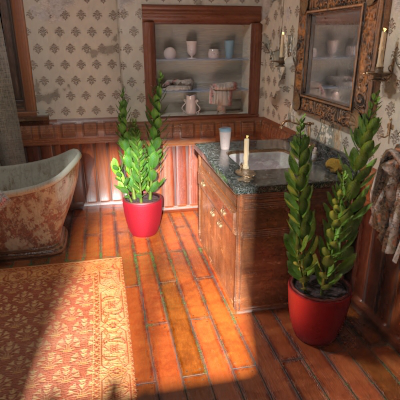}}
\subfloat{
		\includegraphics[scale=0.185]{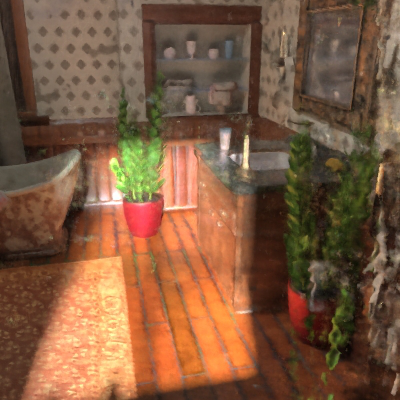}}
\subfloat{
		\includegraphics[scale=0.185]{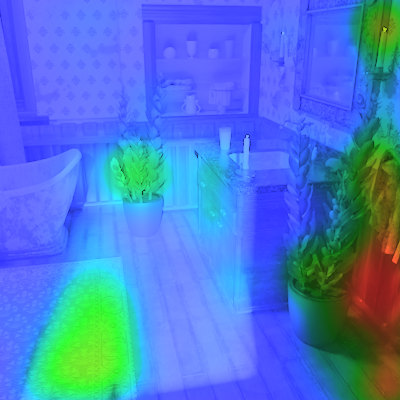}}
\subfloat{
		\includegraphics[scale=0.185]{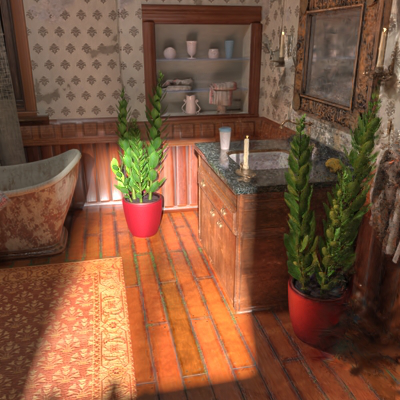}}
\subfloat{
		\includegraphics[scale=0.185]{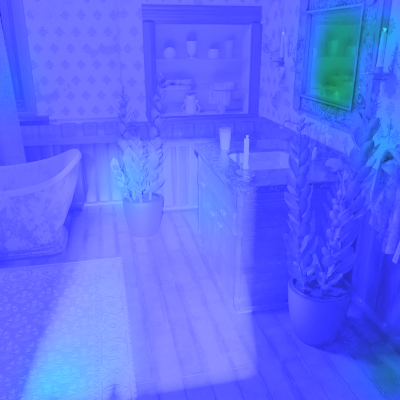}}

\rotatebox{90}{\scriptsize{~~~~~~~~~~~~~chair}}
\subfloat{
		\includegraphics[scale=0.185]{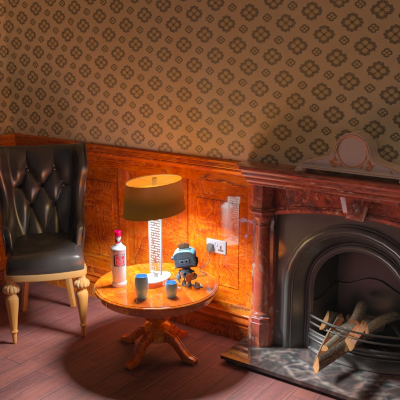}}
\subfloat{
		\includegraphics[scale=0.185]{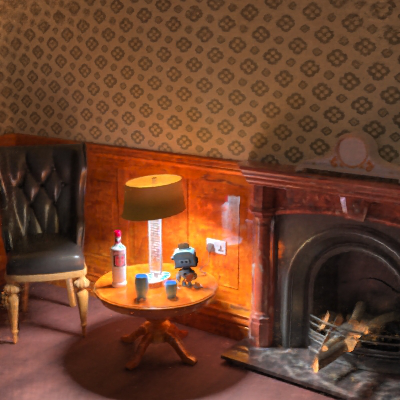}}
\subfloat{
		\includegraphics[scale=0.185]{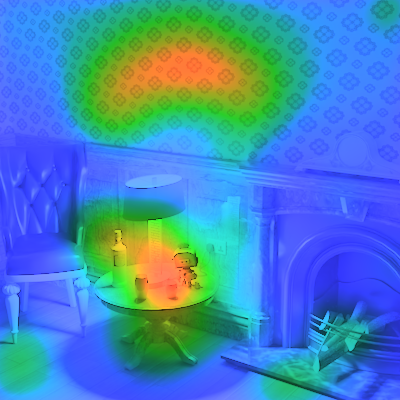}}
\subfloat{
		\includegraphics[scale=0.185]{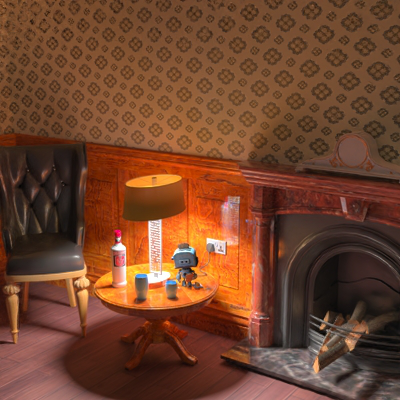}}
\subfloat{
		\includegraphics[scale=0.185]{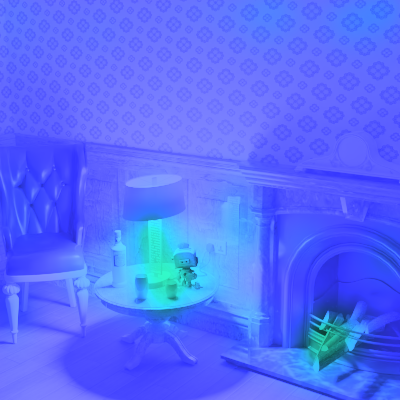}}

\setcounter{subfigure}{0}  

\rotatebox{90}{\scriptsize{~~~~~~~~~~~~~sofa}}
\subfloat[]{
		\includegraphics[scale=0.185]{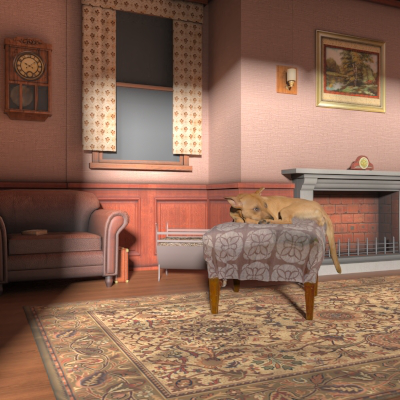}}
\subfloat[]{
		\includegraphics[scale=0.185]{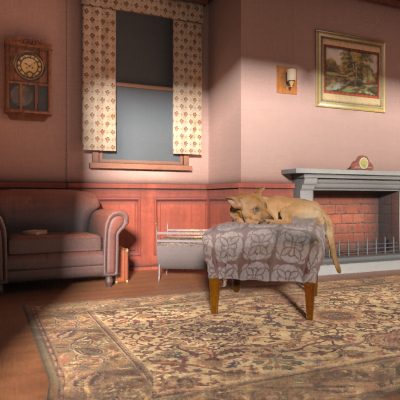}}
\subfloat[]{
		\includegraphics[scale=0.185]{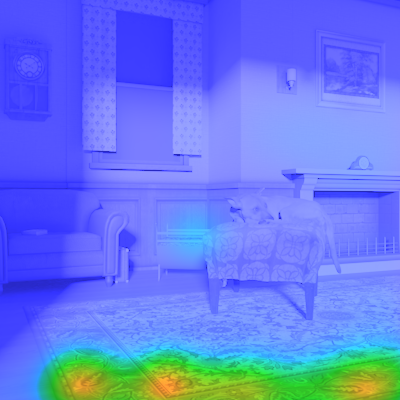}}
\subfloat[]{
		\includegraphics[scale=0.185]{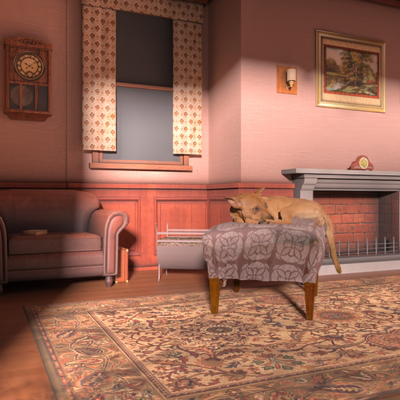}}
\subfloat[]{
		\includegraphics[scale=0.185]{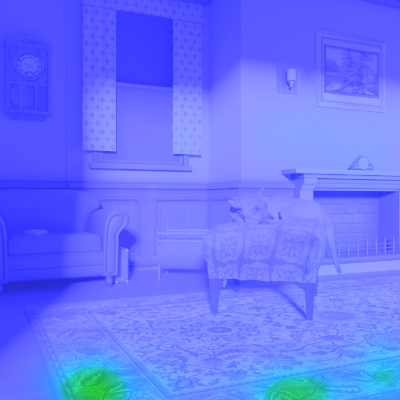}}

\caption{Comparison of HDR image quality for novel viewpoints on syn scenes. (a) represents the GT HDR images, (b) depicts the HDR images rendered by HDRNeRF\cite{huang2022hdr} , while the right side (c) shows their error maps drawn by HDR-VDP. (d) presents the HDR images rendered by our method, and (e) displays the error maps of ours.}
\label{fig:compair hdr syn}
\end{figure}

In Fig. \ref{fig:compair ldr}, we compare our method with HDRNeRF in rendering images under novel viewpoints and exposures. Zoom-in insets reveal that our approach recovers details more accurately than HDRNeRF \cite{huang2022hdr}. Moreover, in the error map, it is evident that our method exhibits fewer errors overall. The novel HDR views of real and synthetic scenes are respectively presented in Fig. \ref{fig:compair hdr real} and Fig. \ref{fig:compair hdr syn}. In Fig. \ref{fig:compair hdr syn}(c) and (e), it can be observed that our method demonstrates smaller errors in HDR-VDP measurements, suggesting closer proximity to the ground truth. Finally, in the supplementary material, we will discuss how our method exhibits greater stability compared to HDRNeRF, as the implicit representation of MLPs may fail to accurately decouple nonlinear components.

\textbf{Ablation studies.} In Table. \ref{tab:Ablation study}, we conducted ablation experiments on several key modules of our method (with or without unit loss, with or without coarse stage, with or without time scaling, and symmetric or asymmetric grid). The left side of the table presents the methods under investigation, while the right side displays the average test results for LDR-OE and LDR-NE. The second row illustrates that, unlike HDRNeRF \cite{huang2022hdr}, our method can still learn a high quality HDR radiance field and produce high-quality images even without utilizing \(\mathcal{L}_u\) to constrain the value at zero (with the goal of learning a specific tone mapper). The third row highlights that our designed coarse-to-fine strategy significantly enhances the learning of tone mapper based on discrete grids. Without coarse-to-fine, the model is susceptible to falling into local optima or requiring extended convergence times. In the fourth row, we conduct ablation experiments on our proposed time scaling. Due to the discrete exposure range of the dataset, without time scaling, the model tends to overfit to each exposure time, resulting in the generation of low-quality LDR images at novel exposure times. Lastly, we verify the effectiveness of our suggested asymmetric grid, particularly in scenarios with high radiance value ranges and irregular density distributions. The results in Table \ref{tab:Ablation study} demonstrate the success of our approach in enhancing quality.

\section{Conclusion}
To recover 3D HDR radiance fields from 2D multi-exposure LDR images, we propose HDRGS, a real-time rendering-supported method with highly interpretable HDR radiance field reconstruction. We redefine the color of Gaussian points as radiance, implying that each Gaussian point radiates radiance rather than color. Then we construct an uneven asymmetric grid as our tone mapper, mapping irradiance received by each pixel to LDR color values. Additionally, to mitigate the grid’s discrete nature and the tendency to fall into local optima, we introduce a novel coarse-to-fine strategy, effectively accelerating model convergence and enhancing reconstruction quality, making it more robust for various complex exposure scenes. Each of our module designs can be mathematically explained. Experimental results on real and synthetic datasets demonstrate that our method achieves state-of-the-art efficiency and quality in reconstructing both LDR images from arbitrary viewpoints and HDR images from arbitrary viewpoints and exposures. To our knowledge, our method marks the pioneering exploration into the potential of Gaussian splatting for 3D HDR reconstruction. Our model and data code will be fully provided for future research endeavors.

\textbf{Limitation:} Our method crudely simulates the physical imaging process, simplifying some details. For instance, we have not taken into account the influence of aperture size and ISO gain. Due to the limitations of the 3DGS\cite{kerbl20233d} method, our method currently exhibits subpar reconstruction results for scenes containing transparent objects.

\newpage
\printbibliography{}


    




\newpage

\appendix

\section{Appendix / supplemental material}

We will provide more detailed information about our experimental procedures in the supplemental material. In Section \ref{AID}, we will explore additional implementation details. Section \ref{Proof} will outline the derivation process of \( f_H \). Section \ref{Dis} will discuss some of the inherent issues in the HDRNeRF model. Finally, Section \ref{AR} will present additional experimental results.

\subsection{Additional Implementation Details}\label{AID}
\textbf{Pruning Strategy.} We utilize the Ray Contribution-Based Pruning technique described in \cite{niemeyer2024radsplat} as our pruning technique. giving each Gaussian point a score:
\begin{equation}
  h(\mathbf{p}_i) = \max_{I_f \in \mathcal{I}_f, r \in I_f} \alpha_i^r\tau_i^r 
\end{equation}

where \( \alpha_i^r\tau_i^r \) represents the contribution of Gaussian point \( \mathbf{p}_i \) to pixel \( r \). A Gaussian point \(\mathbf{p}_i\) can be retained as long as its contribution to any pixel point during a pruning interval exceeds the threshold; otherwise, it must be pruned. This strategy allows for the removal of points that make relatively low contributions to rendering the image. In this case, the threshold is set at 0.02, and pruning begins after 500 rounds, with intervals of 200 rounds each.

It is worth noting  that this pruning technique must be staggered with transparency reset, as it will cause all Gaussian scores \( h(\mathbf p_i) \) to drop below the cutoff value.

\subsection{Proof}\label{Proof}

The functional relationship \(f_H(\cdot)\) between the learnable parameters \(E'\) and the HDR image pixel values \(E\) will be determined here. If no time scaling is performed, then \(E\) satisfies the following equation:
\begin{equation}
C_1 = g_1 (\ln E + \ln t)
\end{equation}
If time scaling is performed, then \(E'\) satisfies the following equation:
\begin{equation}
C_2 = g_2(r \cdot \ln t + s + E')
\end{equation}
Where \(g_1\) and \(g_2\) represent different CRF functions, both of which are monotonic and differentiable. Since the same LDR image is used for supervision regardless of whether time scaling is performed or not, \(C_1=C_2\). Therefore, \(E'\) and \(E\) satisfy:
\begin{equation}
g_1 (\ln t + \ln E) = g_2(r \cdot \ln t + f(\ln E)) ~~~ \\
where ~~ f(\ln E) = E' + s
\end{equation}
Fixing \(E\) , differentiate both sides \textit{w.r.t.} \(\ln t\), we obtain:
\begin{equation}\label{g1}
g_1' = g_2' \cdot r
\end{equation}
Fixing \(t\), differentiate both sides \textit{w.r.t.} \(\ln E\), we obtain:
\begin{equation}\label{g2}
g_1' = g_2' \cdot f'
\end{equation}

From the eq \ref{g1} \ref{g2}, we can infer that \(f' = r\), which implies:
\begin{equation}
\frac {d(E'+s)}{d(\ln E )} = r
\end{equation}
If the unit exposure loss is adopted, then there exists only a scaling relationship between \(g_1\) and \(g_2\), without any offset. Therefore, we can easily derive the final equation:
\begin{equation}
\label{eq:final fh}
E = e^{\frac {(E'+s)}{r}}
\end{equation}

From Eq.\ref{eq:final fh}, we can render HDR images after the model training is completed.

\begin{figure}[!t]
\centering
\subfloat[]{
		\includegraphics[scale=0.16]{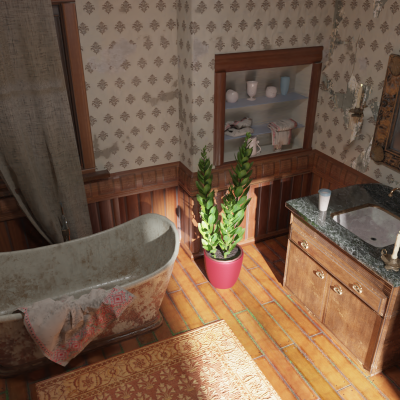}}
\subfloat[]{
		\includegraphics[scale=0.16]{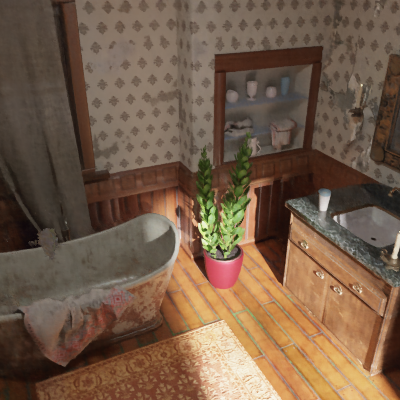}}
\subfloat[]{
		\includegraphics[scale=0.16]{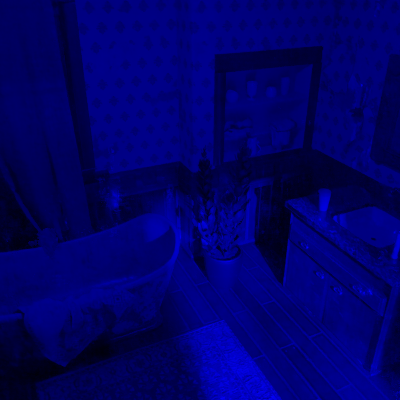}}
\subfloat[]{
		\includegraphics[scale=0.16]{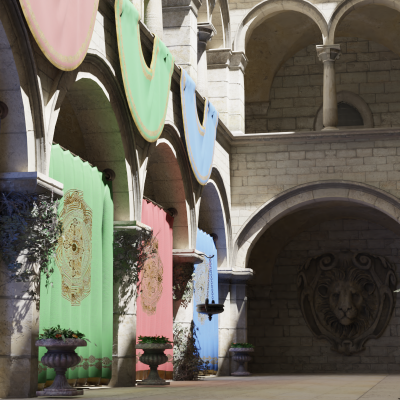}}
\subfloat[]{
		\includegraphics[scale=0.16]{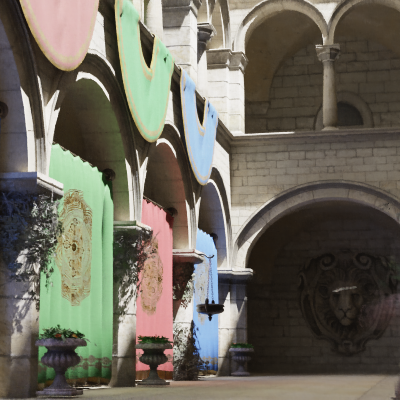}}
\subfloat[]{
		\includegraphics[scale=0.16]{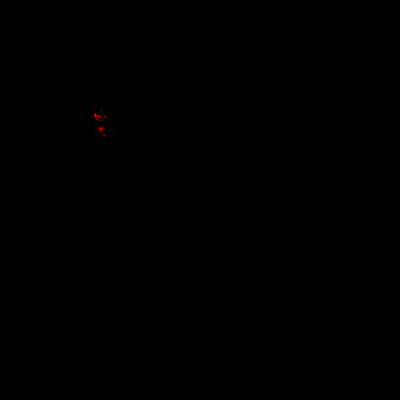}}
\caption{\textbf{Failure cases.} (a) and (d) are the GT images, (b) and (e) are LDR images rendered by HDRNeRF\cite{huang2022hdr} , while (c) and (f) are HDR images.} 
\label{fig: discuss}
\end{figure}

\subsection{Discussion}\label{Dis}
During our experiments, we observed instances where the tone mapper of HDRNeRF \cite{huang2022hdr} occasionally overfits to LDR, leading to an inaccurate HDR radiance field. As depicted in Fig. \ref{fig: discuss}, it's evident that directly utilizing MLP in HDRNeRF to model the tone mapper is unstable. Employing three separate MLPs in HDRNeRF \cite{huang2022hdr} to handle the RGB channels independently might induce crosstalk problems, where interference between channels could potentially lead to being trapped in local optimal solutions. While HDRNeRF \cite{huang2022hdr}  learns the correct LDR, errors arise in HDR radiance field construction. Furthermore, in \cite{jun2022hdr,wu2024fast}, the authors mention that constructing a tone mapping module based on MLPs may fail to correctly decouple nonlinear components.

\subsection{Additional Results}\label{AR}
\begin{figure}
\centering
\rotatebox{90}{\scriptsize{~~~~~~~~~~~~~GT}}\hspace{0.1cm}
\subfloat{
		\includegraphics[scale=0.185]{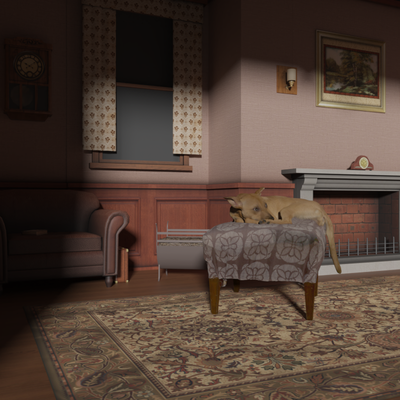}}
\subfloat{
		\includegraphics[scale=0.185]{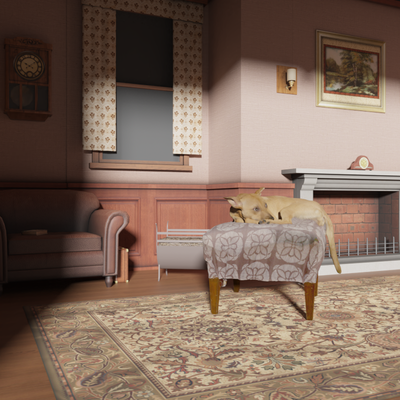}}
\subfloat{
		\includegraphics[scale=0.185]{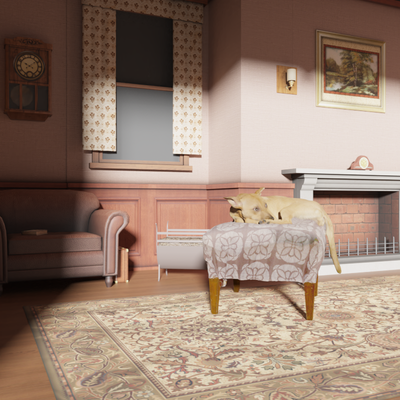}}
\subfloat{
		\includegraphics[scale=0.185]{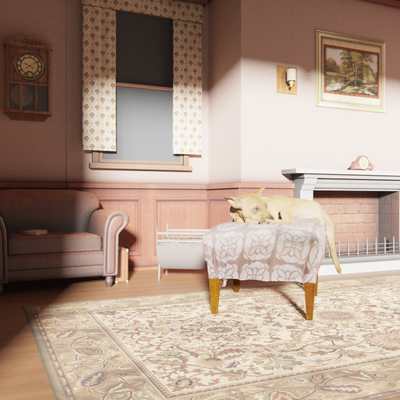}}
\subfloat{
		\includegraphics[scale=0.185]{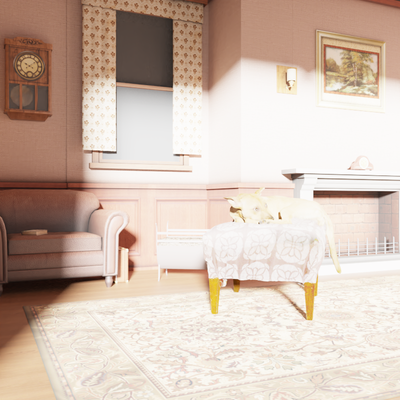}}

\rotatebox{90}{\scriptsize{~~~~~~~~~~~~~3DGS}}\hspace{0.1cm}
\subfloat{
		\includegraphics[scale=0.185]{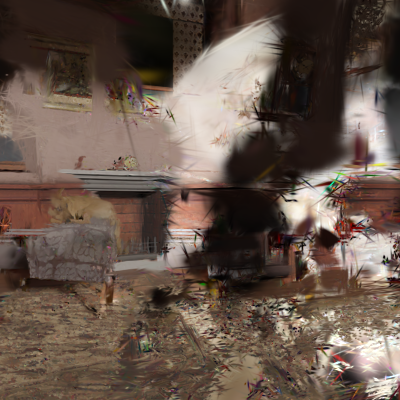}}
\subfloat{
		\includegraphics[scale=0.185]{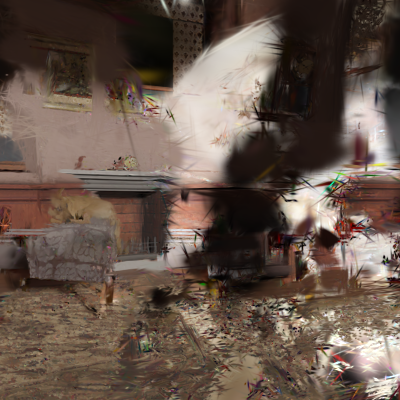}}
\subfloat{
		\includegraphics[scale=0.185]{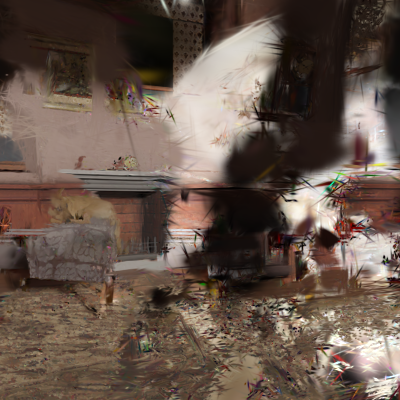}}
\subfloat{
		\includegraphics[scale=0.185]{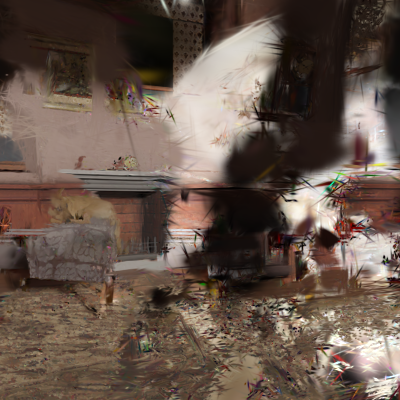}}
\subfloat{
		\includegraphics[scale=0.185]{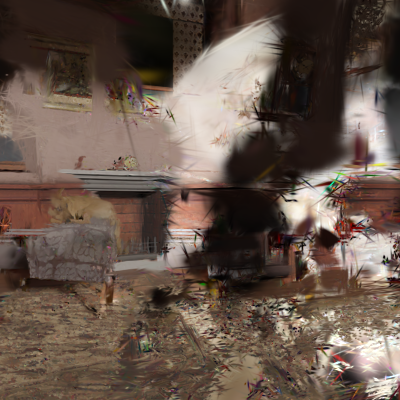}}

\rotatebox{90}{\scriptsize{~~~~~~~~~~~~~HDRNeRF}}\hspace{0.1cm}
\subfloat{
		\includegraphics[scale=0.185]{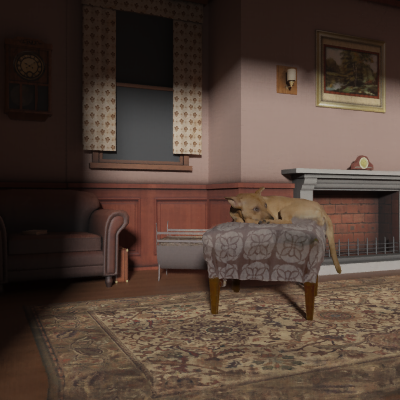}}
\subfloat{
		\includegraphics[scale=0.185]{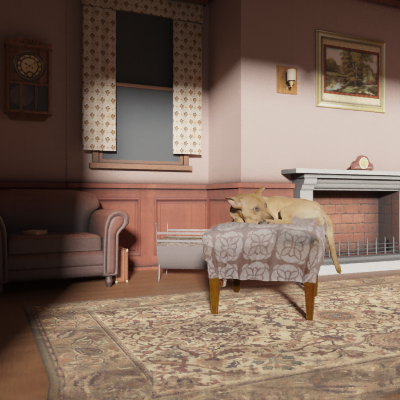}}
\subfloat{
		\includegraphics[scale=0.185]{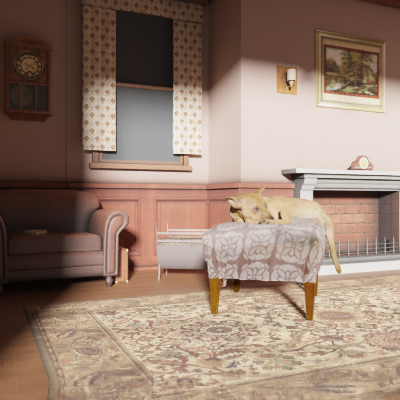}}
\subfloat{
		\includegraphics[scale=0.185]{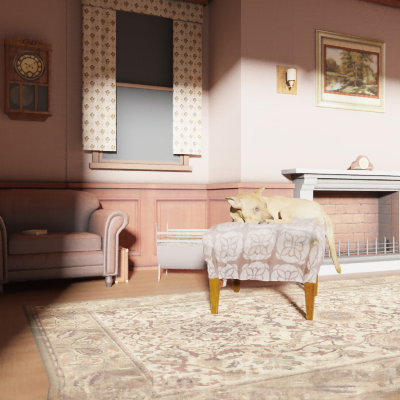}}
\subfloat{
		\includegraphics[scale=0.185]{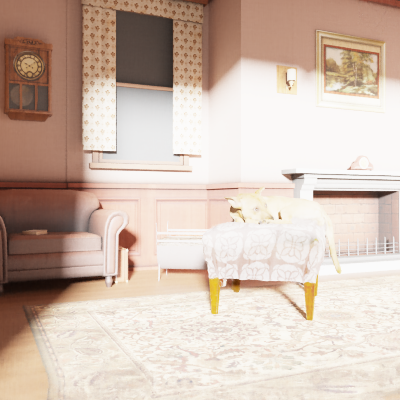}}
  
\rotatebox{90}{\scriptsize{~~~~~~~~~~~~~Ours}}\hspace{0.1cm}
\subfloat{
		\includegraphics[scale=0.185]{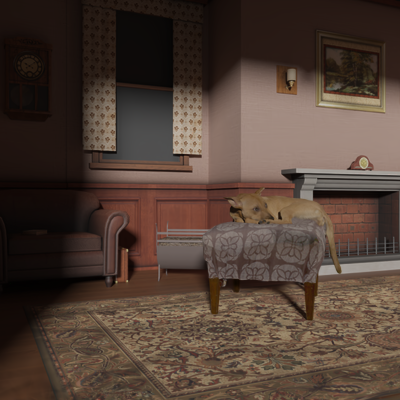}}
\subfloat{
		\includegraphics[scale=0.185]{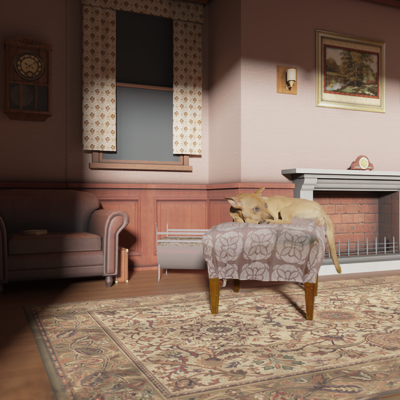}}
\subfloat{
		\includegraphics[scale=0.185]{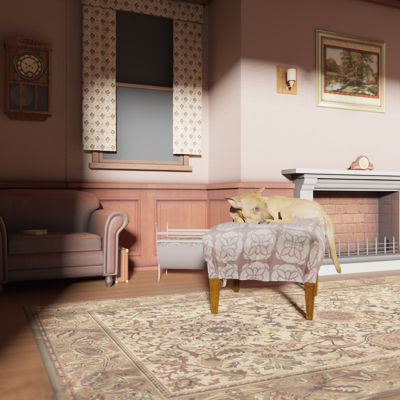}}
\subfloat{
		\includegraphics[scale=0.185]{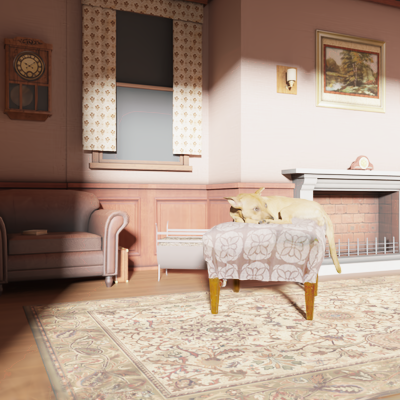}}
\subfloat{
		\includegraphics[scale=0.185]{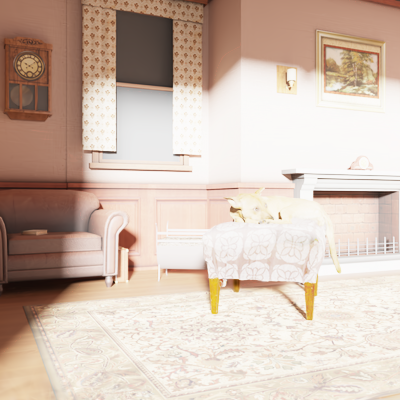}}

\rotatebox{90}{\scriptsize{~~~~~~~~~~~~~Unit exp loss}}\hspace{0.1cm}
\subfloat{
		\includegraphics[scale=0.185]{figs/ablation/gt/75.png}}
\subfloat{
		\includegraphics[scale=0.185]{figs/ablation/gt/76.png}}
\subfloat{
		\includegraphics[scale=0.185]{figs/ablation/gt/77.png}}
\subfloat{
		\includegraphics[scale=0.185]{figs/ablation/gt/78.png}}
\subfloat{
		\includegraphics[scale=0.185]{figs/ablation/gt/79.png}}

\rotatebox{90}{\scriptsize{~~~~~~~~~~~~~Coarse stage}}\hspace{0.1cm}
\subfloat{
		\includegraphics[scale=0.185]{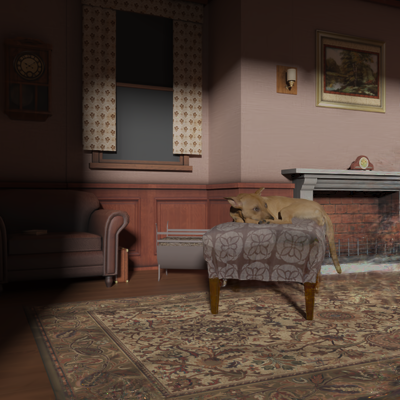}}
\subfloat{
		\includegraphics[scale=0.185]{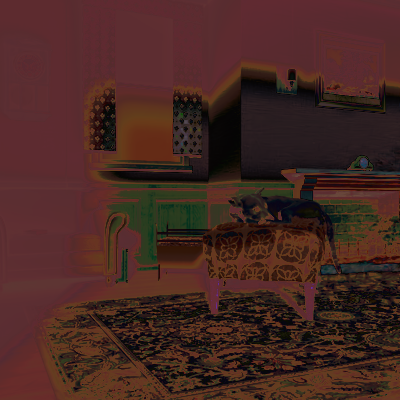}}
\subfloat{
		\includegraphics[scale=0.185]{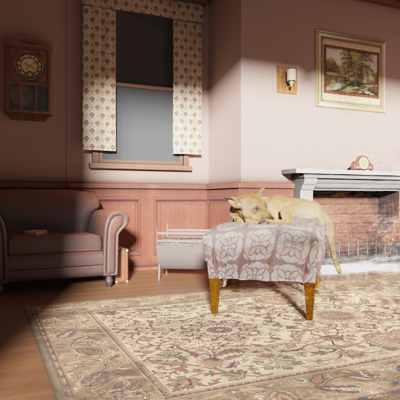}}
\subfloat{
		\includegraphics[scale=0.185]{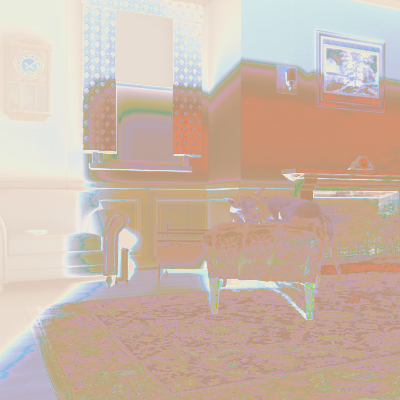}}
\subfloat{
		\includegraphics[scale=0.185]{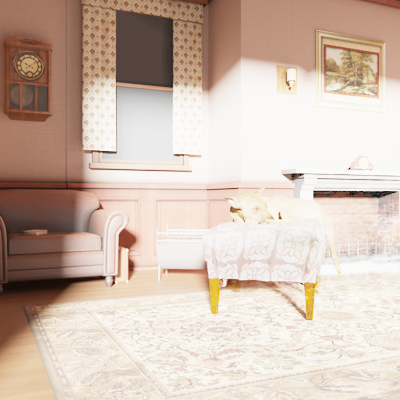}}

\rotatebox{90}{\scriptsize{~~~~~~~~~~~~~Time scaling}}\hspace{0.1cm}
\subfloat{
		\includegraphics[scale=0.185]{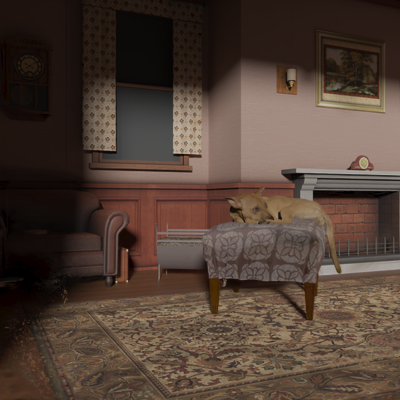}}
\subfloat{
		\includegraphics[scale=0.185]{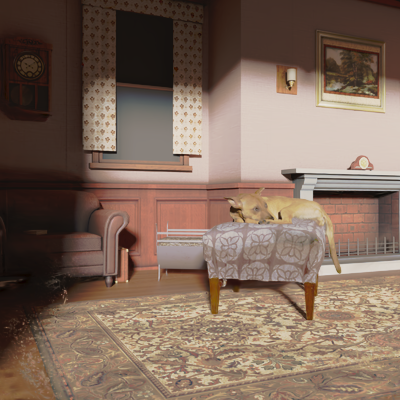}}
\subfloat{
		\includegraphics[scale=0.185]{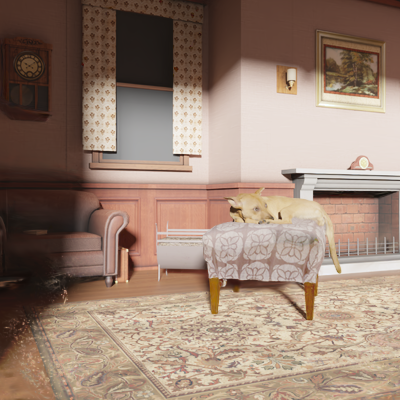}}
\subfloat{
		\includegraphics[scale=0.185]{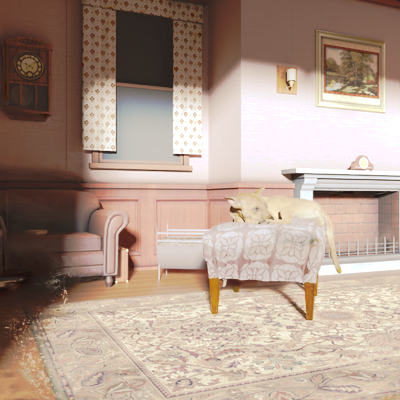}}
\subfloat{
		\includegraphics[scale=0.185]{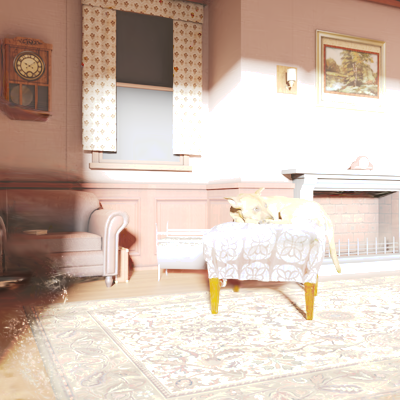}}

\setcounter{subfigure}{0}  
\rotatebox{90}{\scriptsize{~~~~~~~~~~~~~Symmetric}}\hspace{0.1cm}
\subfloat[\(t_1\)]{
		\includegraphics[scale=0.185]{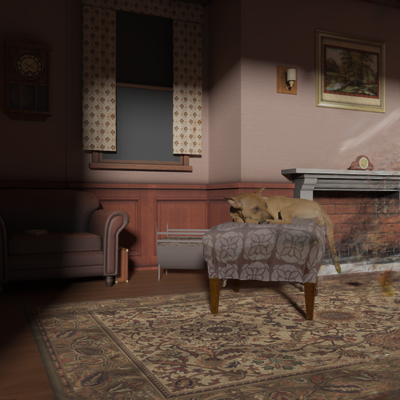}}
\subfloat[\(t_2\)]{
		\includegraphics[scale=0.185]{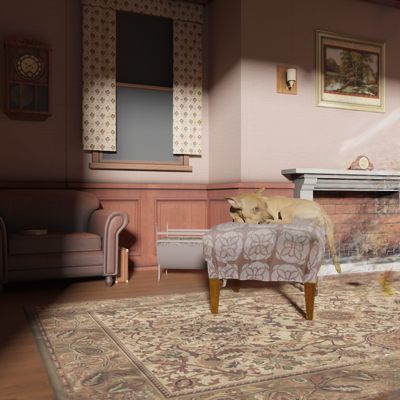}}
\subfloat[\(t_3\)]{
		\includegraphics[scale=0.185]{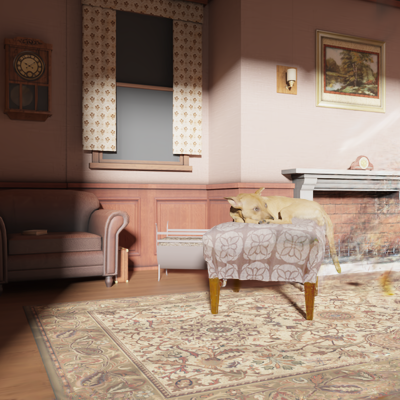}}
\subfloat[\(t_4\)]{
		\includegraphics[scale=0.185]{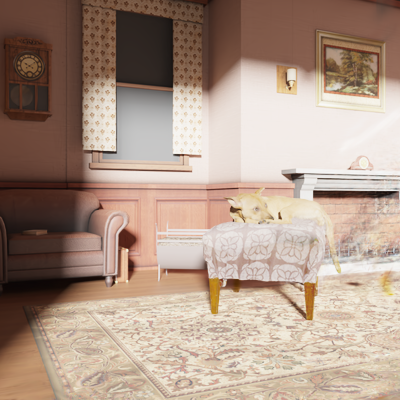}}
\subfloat[\(t_5\)]{
		\includegraphics[scale=0.185]{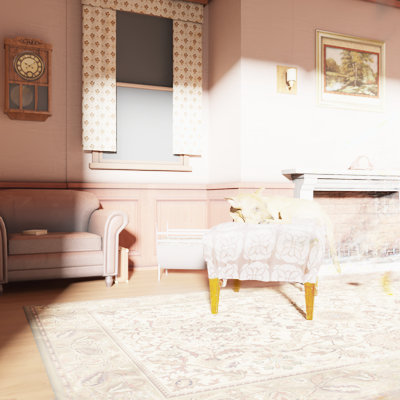}}

\caption{ Comparison of LDR image quality under exposure time \(\{t_1,t_2,t_3,t_4,t_5\}\) for novel viewpoints on "sofa". The exposure time of training set is randomly selected from  \(\{t_1,t_3,t_5\}\).} 
\label{fig:ablation}
\end{figure}

\begin{figure}
\centering
\rotatebox{90}{\scriptsize{~~~~~~~~~~~~~bathroom}}\hspace{0.1cm}
\subfloat{
		\includegraphics[scale=0.15]{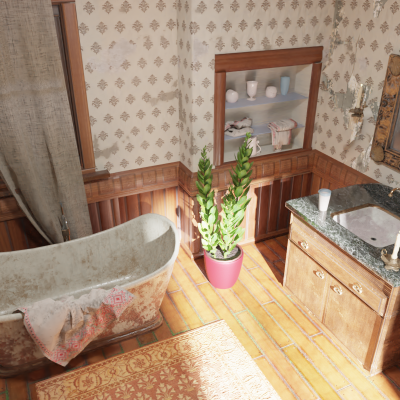}}
\subfloat{
		\includegraphics[scale=0.15]{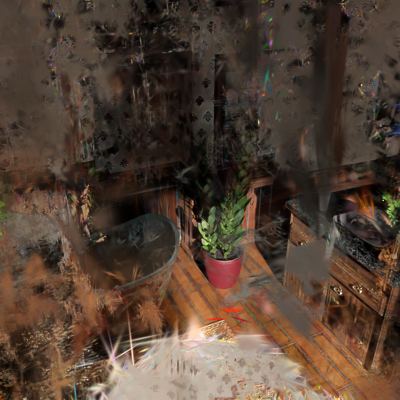}}
\subfloat{
		\includegraphics[scale=0.15]{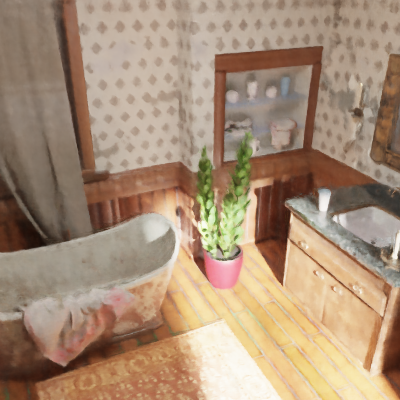}}
\subfloat{
		\includegraphics[scale=0.15]{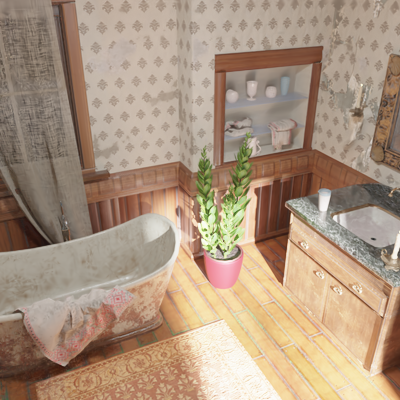}}
\subfloat{
		\includegraphics[scale=0.15]{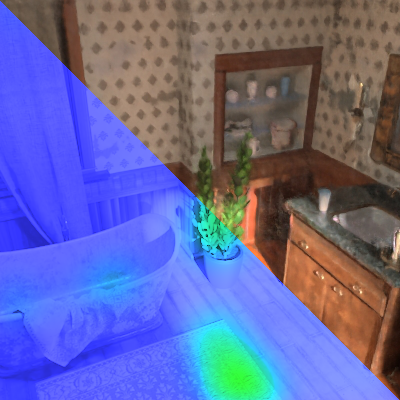}}
\subfloat{
		\includegraphics[scale=0.15]{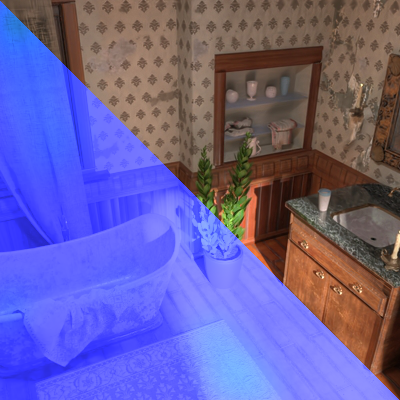}}
  
\rotatebox{90}{\scriptsize{~~~~~~~~~~~~~bear}}\hspace{0.1cm}
\subfloat{
		\includegraphics[scale=0.15]{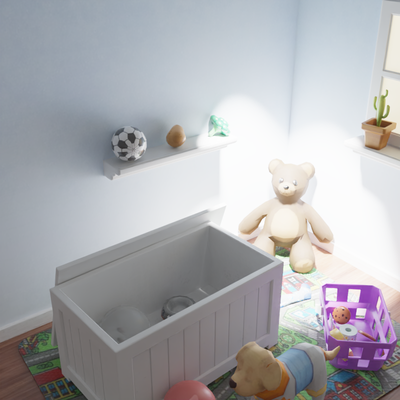}}
\subfloat{
		\includegraphics[scale=0.15]{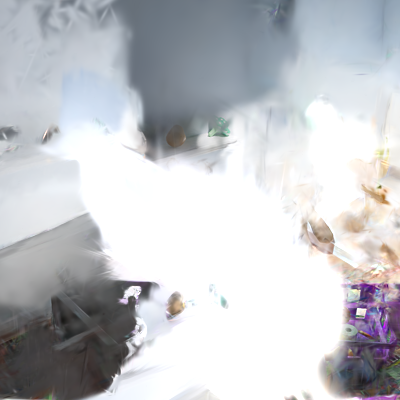}}
\subfloat{
		\includegraphics[scale=0.15]{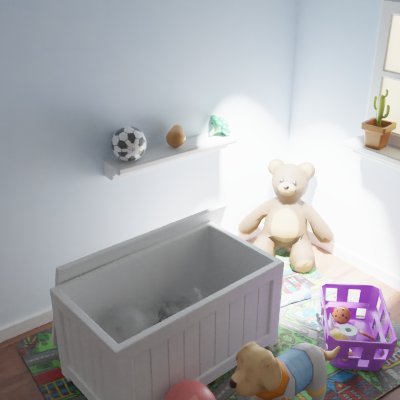}}
\subfloat{
		\includegraphics[scale=0.15]{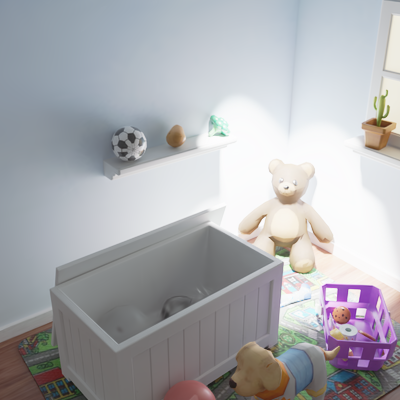}}
\subfloat{
		\includegraphics[scale=0.15]{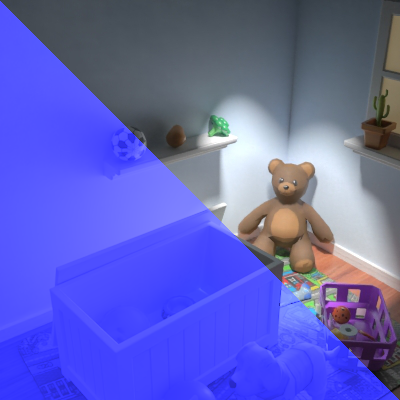}}
\subfloat{
		\includegraphics[scale=0.15]{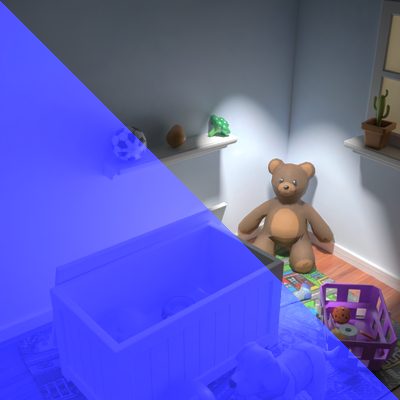}}

\rotatebox{90}{\scriptsize{~~~~~~~~~~~~~chair}}\hspace{0.1cm}
\subfloat{
		\includegraphics[scale=0.15]{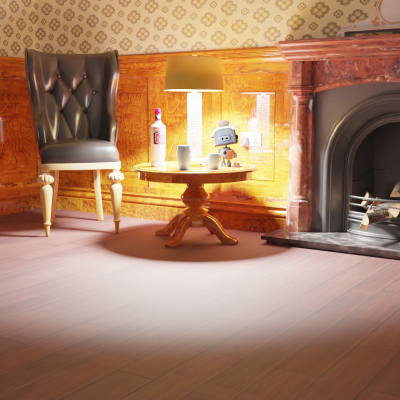}}
\subfloat{
		\includegraphics[scale=0.15]{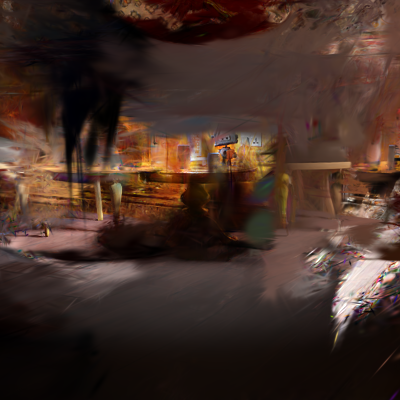}}
\subfloat{
		\includegraphics[scale=0.15]{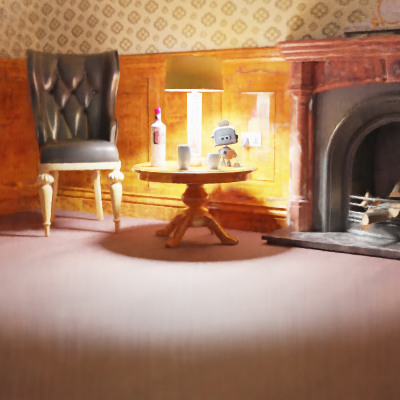}}
\subfloat{
		\includegraphics[scale=0.15]{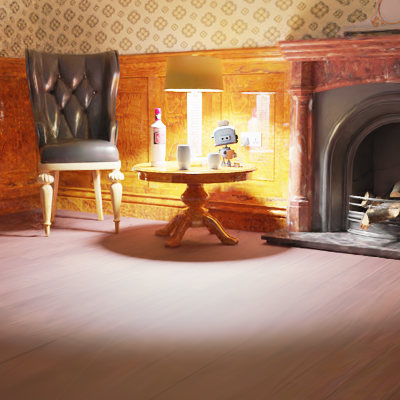}}
\subfloat{
		\includegraphics[scale=0.15]{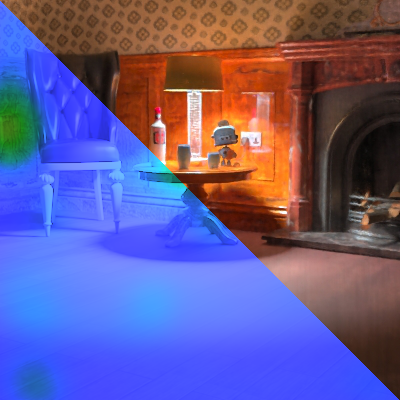}}
\subfloat{
		\includegraphics[scale=0.15]{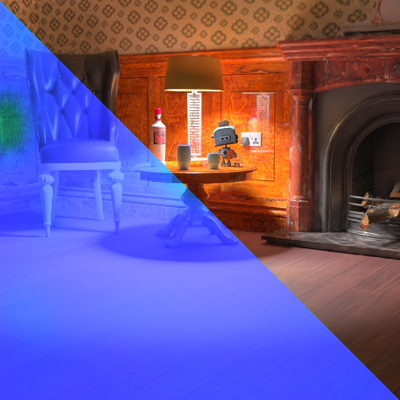}}

\rotatebox{90}{\scriptsize{~~~~~~~~~~~~~desk}}\hspace{0.1cm}
\subfloat{
		\includegraphics[scale=0.15]{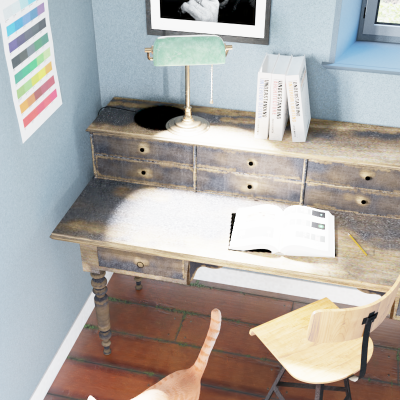}}
\subfloat{
		\includegraphics[scale=0.15]{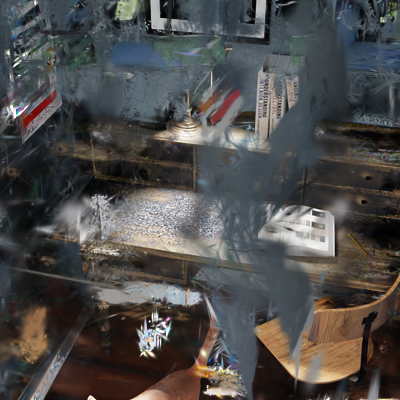}}
\subfloat{
		\includegraphics[scale=0.15]{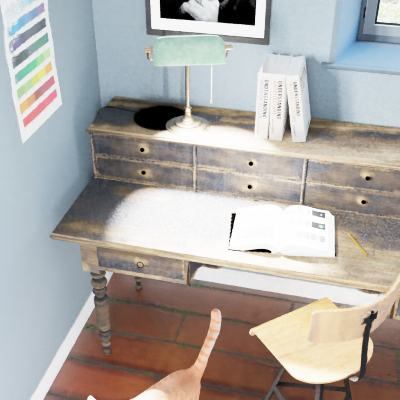}}
\subfloat{
		\includegraphics[scale=0.15]{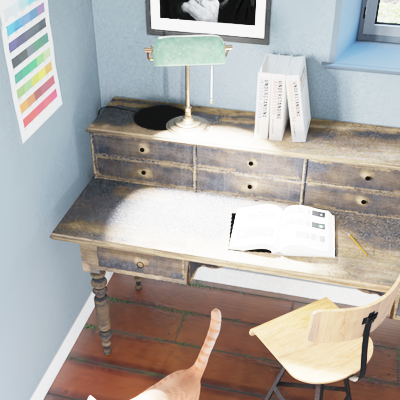}}
\subfloat{
		\includegraphics[scale=0.15]{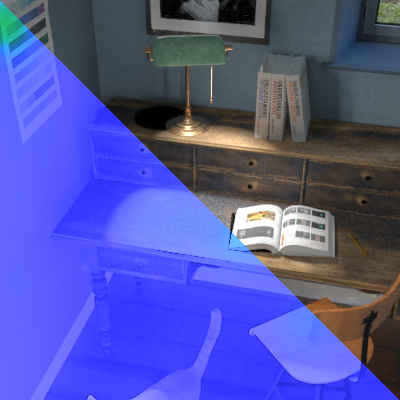}}
\subfloat{
		\includegraphics[scale=0.15]{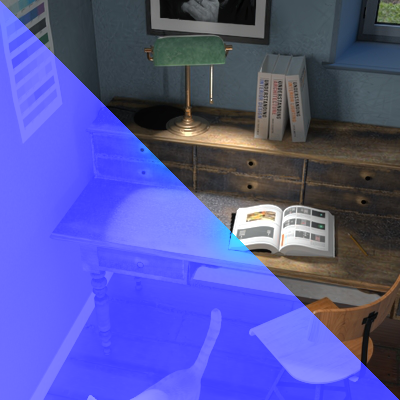}}

\rotatebox{90}{\scriptsize{~~~~~~~~~~~~~diningroom}}\hspace{0.1cm}
\subfloat{
		\includegraphics[scale=0.15]{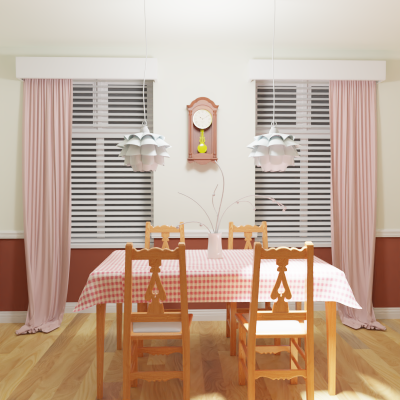}}
\subfloat{
		\includegraphics[scale=0.15]{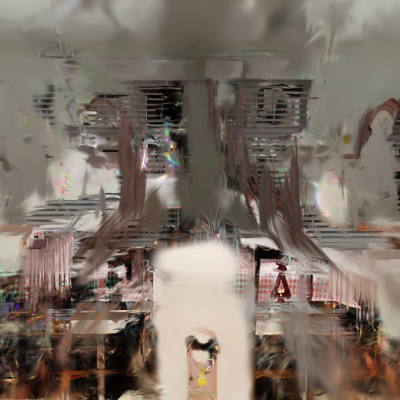}}
\subfloat{
		\includegraphics[scale=0.15]{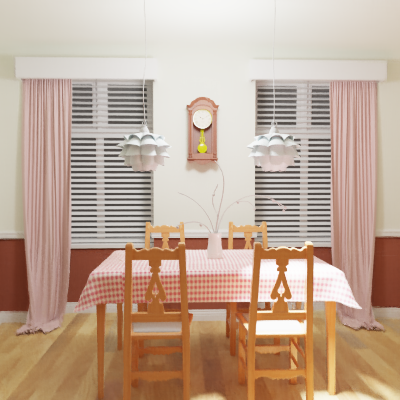}}
\subfloat{
		\includegraphics[scale=0.15]{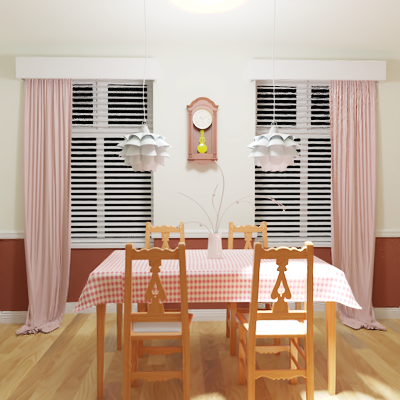}}
\subfloat{
		\includegraphics[scale=0.15]{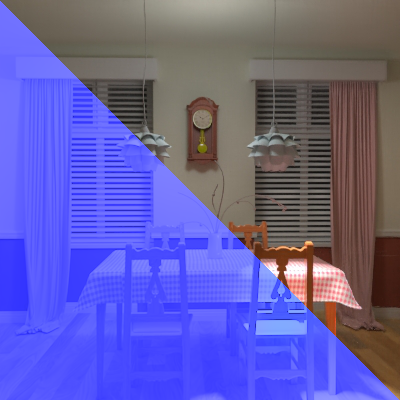}}
\subfloat{
		\includegraphics[scale=0.15]{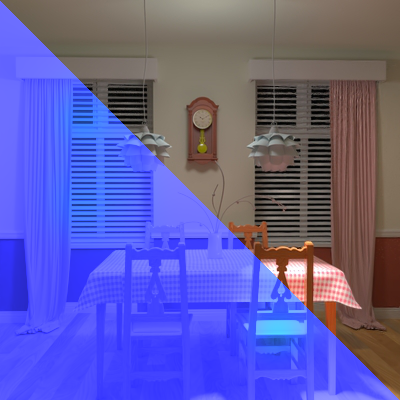}}

\rotatebox{90}{\scriptsize{~~~~~~~~~~~~~dog}}\hspace{0.1cm}
\subfloat{
		\includegraphics[scale=0.15]{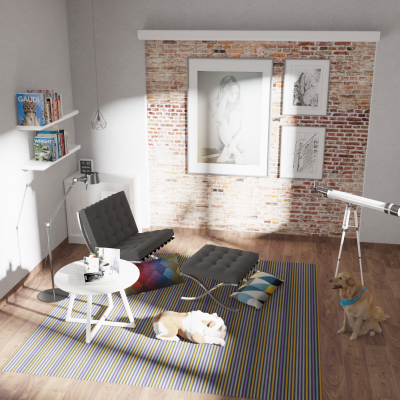}}
\subfloat{
		\includegraphics[scale=0.15]{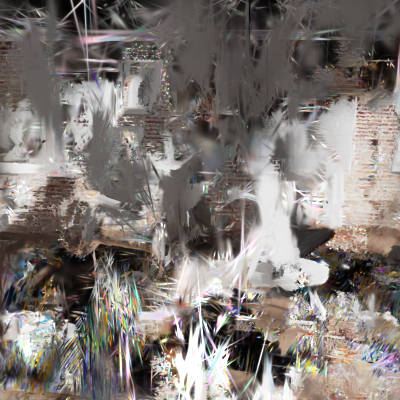}}
\subfloat{
		\includegraphics[scale=0.15]{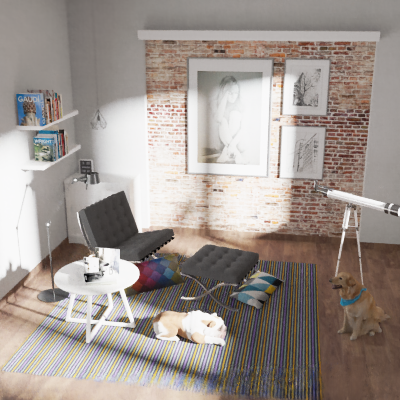}}
\subfloat{
		\includegraphics[scale=0.15]{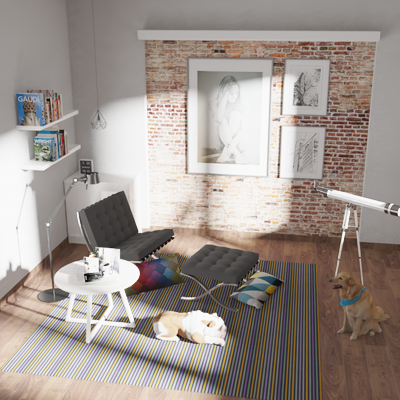}}
\subfloat{
		\includegraphics[scale=0.15]{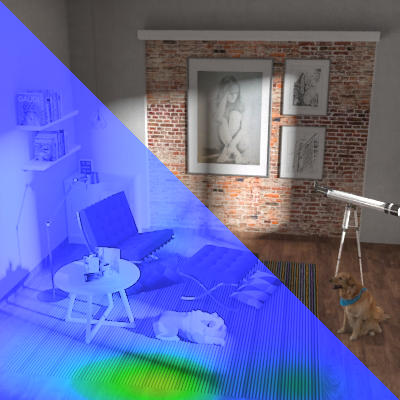}}
\subfloat{
		\includegraphics[scale=0.15]{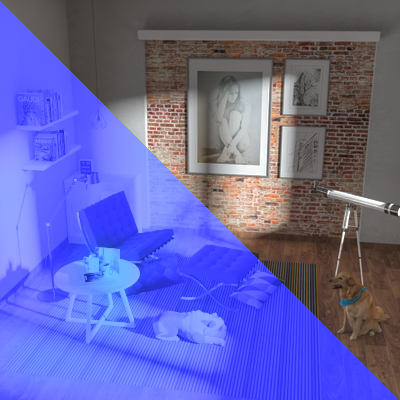}}

\rotatebox{90}{\scriptsize{~~~~~~~~~~~~~sofa}}\hspace{0.1cm}
\subfloat{
		\includegraphics[scale=0.15]{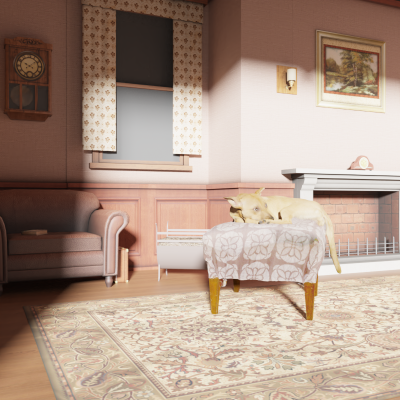}}
\subfloat{
		\includegraphics[scale=0.15]{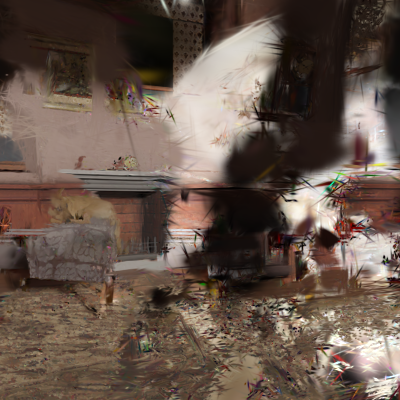}}
\subfloat{
		\includegraphics[scale=0.15]{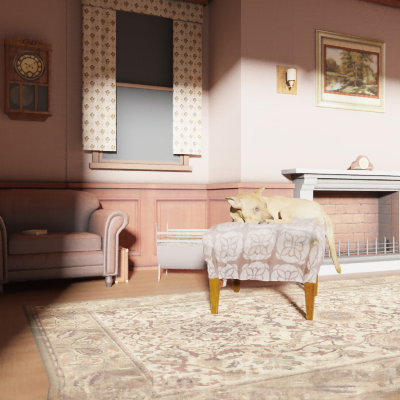}}
\subfloat{
		\includegraphics[scale=0.15]{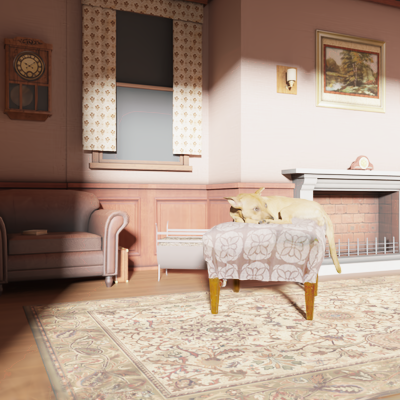}}
\subfloat{
		\includegraphics[scale=0.15]{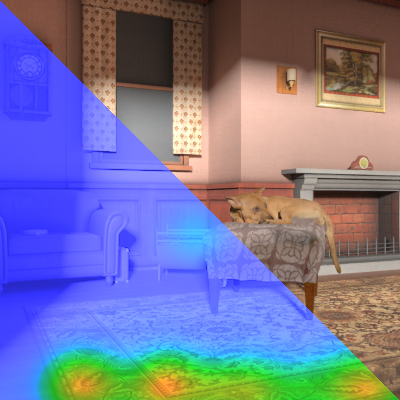}}
\subfloat{
		\includegraphics[scale=0.15]{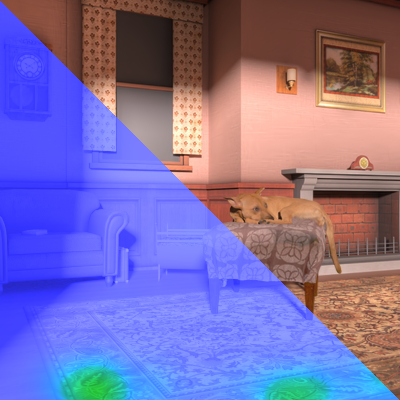}}
  
\setcounter{subfigure}{0}  
\rotatebox{90}{\scriptsize{~~~~~~~~~~~~~sponza}}\hspace{0.1cm}
\subfloat[GT]{
		\includegraphics[scale=0.15]{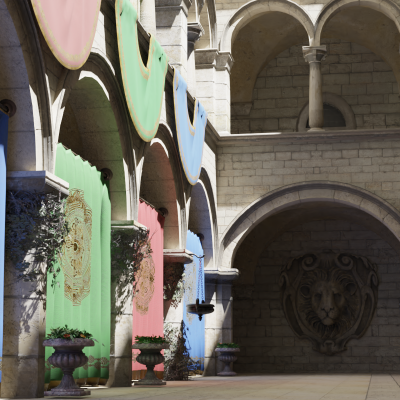}}
\subfloat[3DGS]{
		\includegraphics[scale=0.15]{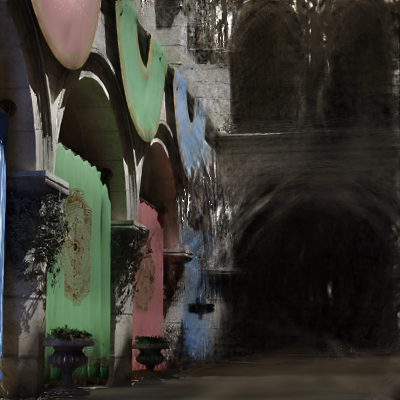}}
\subfloat[HDR-NeRF]{
		\includegraphics[scale=0.15]{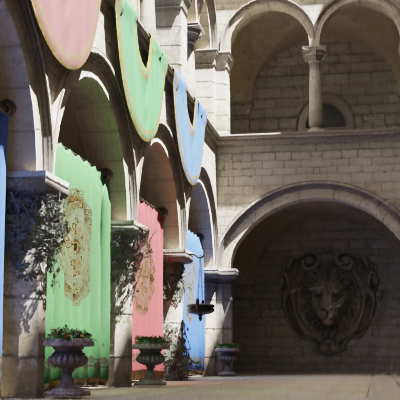}}
\subfloat[Ours]{
		\includegraphics[scale=0.15]{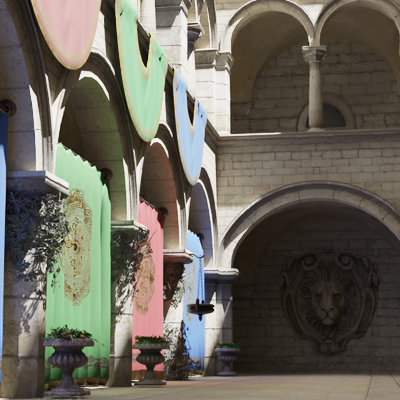}}
\subfloat[HDR-hdrnerf]{
		\includegraphics[scale=0.15]{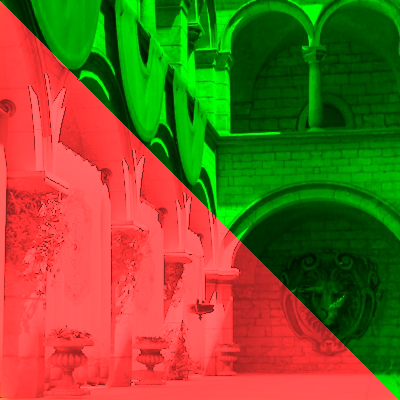}}
\subfloat[HDR-Ours]{
		\includegraphics[scale=0.15]{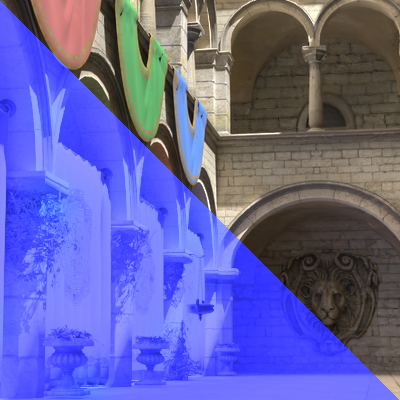}}
  
\caption{Results of the synthesis datasets. The first four columns represent LDR images, while the last two columns depict HDR images and the error maps generated by HDR-VDP.} 
\label{fig:all compair}
\end{figure}

\begin{figure}
\centering
\subfloat[bathroom]{
		\includegraphics[scale=0.2]{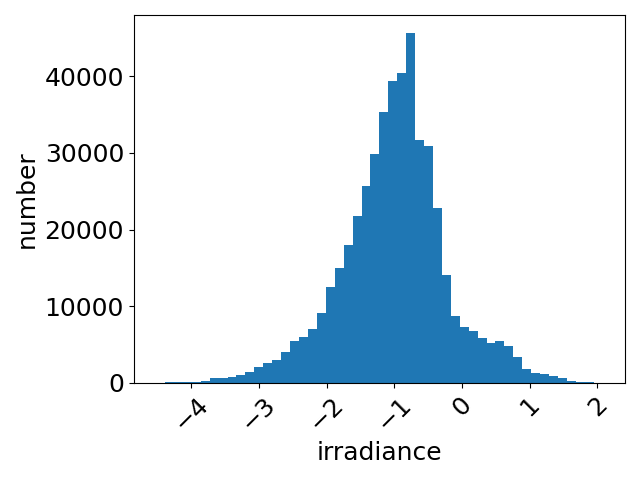}}
\subfloat[bear]{
		\includegraphics[scale=0.2]{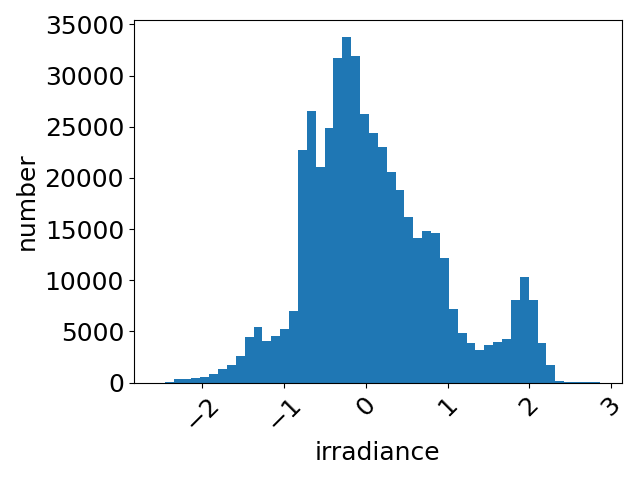}}
\subfloat[chair]{
		\includegraphics[scale=0.2]{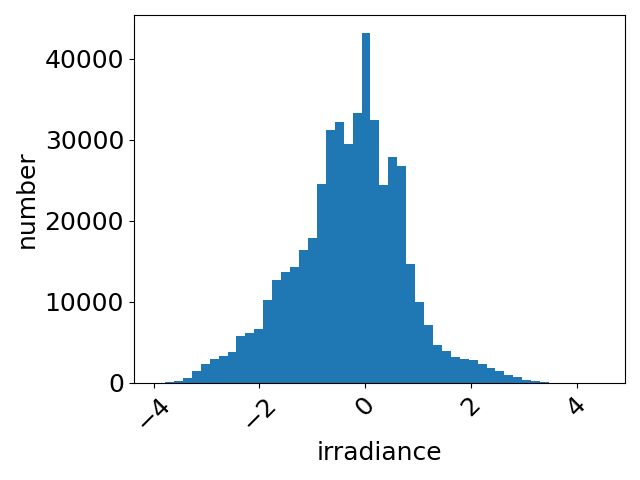}}
\subfloat[desk]{
		\includegraphics[scale=0.2]{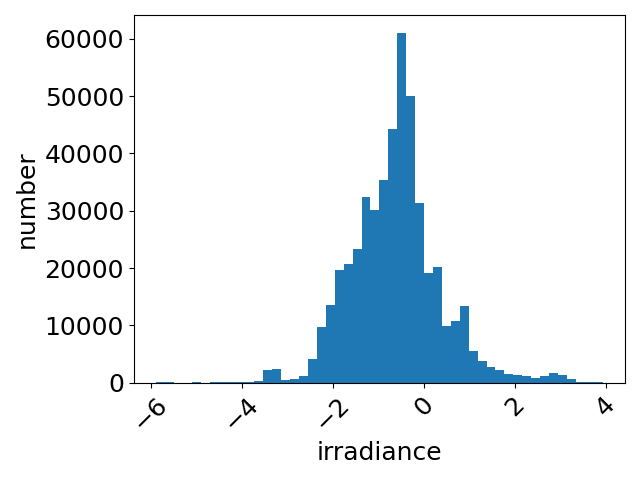}}

\subfloat[diningroom]{
		\includegraphics[scale=0.2]{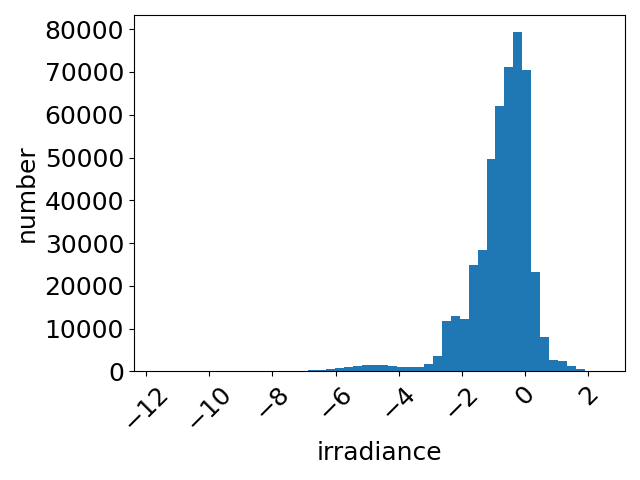}}
\subfloat[dog]{
		\includegraphics[scale=0.2]{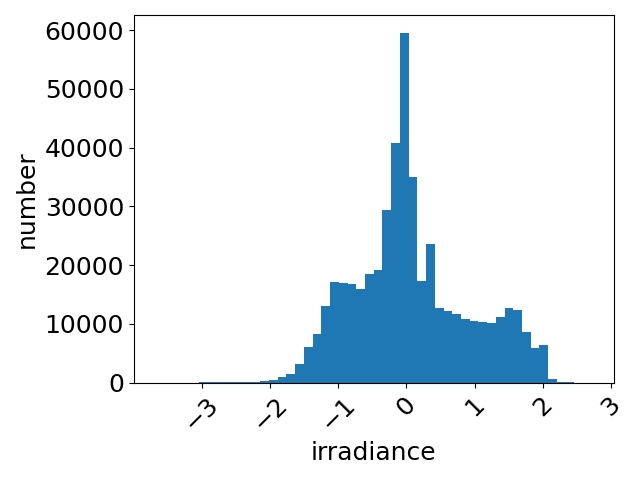}}
\subfloat[sofa]{
		\includegraphics[scale=0.2]{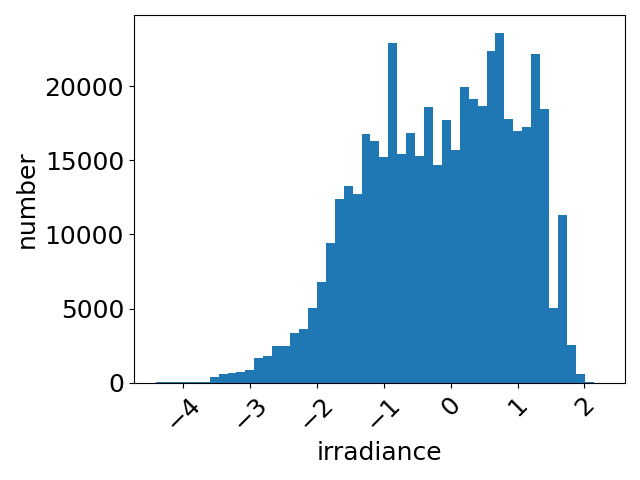}}
\subfloat[sponza]{
		\includegraphics[scale=0.2]{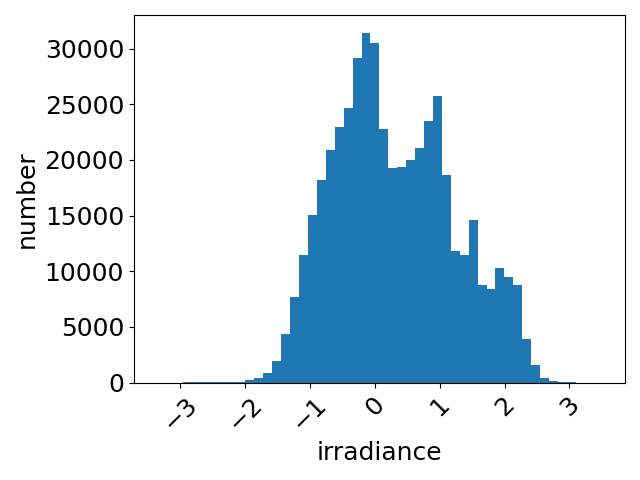}}

\subfloat[box]{
		\includegraphics[scale=0.2]{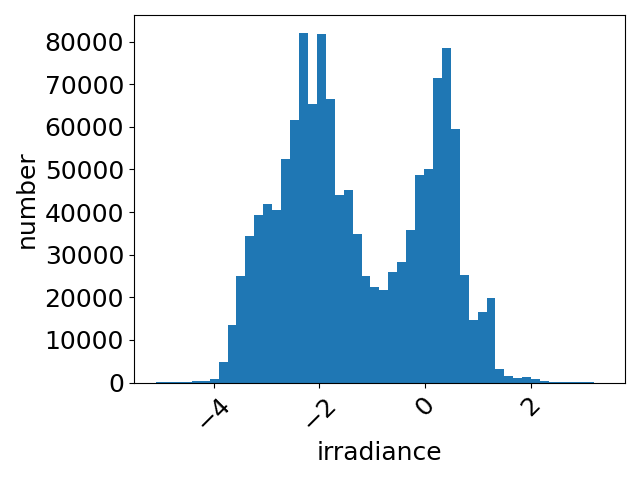}}
\subfloat[computer]{
		\includegraphics[scale=0.2]{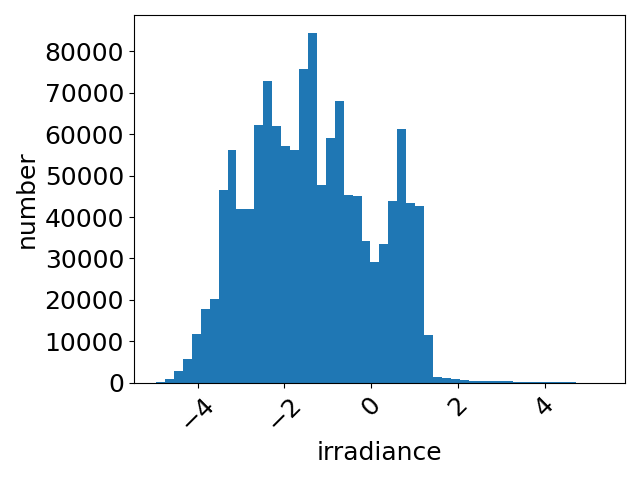}}
\subfloat[flower]{
		\includegraphics[scale=0.2]{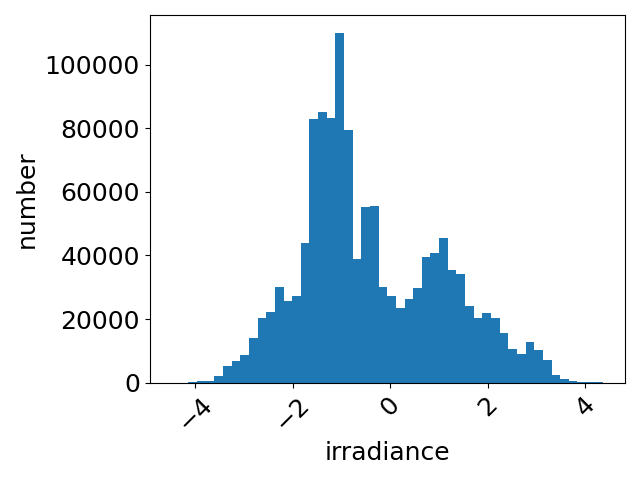}}
\subfloat[luckycat]{
		\includegraphics[scale=0.2]{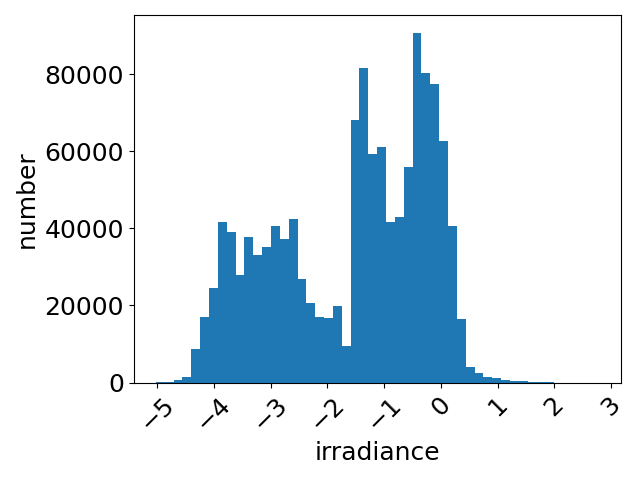}}
  
\caption{The distribution of irradiance.} 
\label{fig:irradiance}
\end{figure}

In this section, we present more experimental results. Table. \ref{tab:all hdr} shows the HDR measurement results compared to HDRNeRF on the synthetic dataset. Tab\ref{tab:all syn ldr} and Tab \ref{tab:all real ldr} demonstrate the LDR measurement results of our method compared to other baseline methods across various scenes. In Fig\ref{fig:ablation}, we present LDR images of our various ablation methods (with/without coarse stage, with/without unit exposure loss, with/without t scaling , asymmetric or symmetric grid ) and other methods under the same viewpoint but different exposure time. In the ablation experiments of the coarse stage, we can clearly see that without the coarse stage, the asymmetric grid would easily overfit to the exposure time used in the training dataset. In Fig\ref{fig:irradiance}, we show the distribution of irradiance \(E'\) for each scene in the first frame of the test dataset. It can be observed that the radiance distribution of most scenes is highly uneven, especially scenes like "diningroom", which easily lead the grid into local optimization problems.  In Fig.\ref{fig:all compair}, we present additional results, where (e) and (f) respectively show HDR images rendered by HDR-NeRF and our method. The upper right triangles of these images are the rendered HDR images, while the lower left triangles are error maps drawn by HDR-VDP.

\begin{table}[] 
\centering
    \caption{ LDR image quality assessment, LDR-OE represents the use of LDR images with exposures \(\{t_1,t_3,t_5\}\) as the training set, while LDR-NE represents the use of LDR images with exposures \(\{t_2,t_4\}\) as the training set. The exposure times of the test dataset are \( \{ t_1, t_2, t_3, t_4, t_5 \} \). The \textcolor{red}{best} and the \textcolor{blue}{second best} results are denoted by red and blue.}
    \label{tab:all syn ldr}
\begin{threeparttable}
\resizebox{\textwidth}{!} 
{
\begin{tabular}{c|c|ccccccccccccc}
\toprule[2pt]
\multirow{2}{*}{} & \multirow{2}{*}{} & \multicolumn{3}{c}{Bathroom} & \multicolumn{3}{c}{Bear} & \multicolumn{3}{c}{Chair} & \multicolumn{3}{c}{Diningroom}\\
    &   & \multicolumn{1}{c}{PSNR}\(\uparrow\) & \multicolumn{1}{c}{SSIM} \(\uparrow\) & \multicolumn{1}{c}{LPIPS}\(\downarrow \)
        & \multicolumn{1}{c}{PSNR}\(\uparrow\)& \multicolumn{1}{c}{SSIM} \(\uparrow\) & \multicolumn{1}{c}{LPIPS}\(\downarrow \)
        & \multicolumn{1}{c}{PSNR}\(\uparrow\) & \multicolumn{1}{c}{SSIM} \(\uparrow\) & \multicolumn{1}{c}{LPIPS}\(\downarrow \)
        & \multicolumn{1}{c}{PSNR}\(\uparrow\)& \multicolumn{1}{c}{SSIM} \(\uparrow\) & \multicolumn{1}{c}{LPIPS}\(\downarrow \)
        \\ 
\midrule[1pt]
\multirow{4}{*}{LDR-OE} 
& NeRF  \tnote{*}    & 14.59 & 0.429 & 0.424 & 11.97 & 0.560 & 0.515 & 12.23 & 0.422 & 0.492 & 12.50 & 0.378 & 0.600  \\ 
& HDR-P  \tnote{+}  &  6.73 &	0.348 &	0.669 &10.77 &	0.637 &	0.553& 10.29 &	0.404 &	0.861 &11.34 &	0.439 &	0.670  \\ 
& NeRF-W \tnote{*}   & 29.64  & 0.900  &\textcolor{blue}{0.055} & 32.24 & 0.978 & \textcolor{blue}{0.021} & 28.01 & 0.840 & 0.161  & 32.25 & \textcolor{blue}{0.979} & \textcolor{blue}{0.016} \\ 

& HDR-NeRF  & \textcolor{blue}{32.81} & \textcolor{blue}{0.887} & 0.121 & \textcolor{blue}{42.50} & \textcolor{blue}{0.989} & 0.097 & \textcolor{blue}{32.33} & \textcolor{blue}{0.901} & \textcolor{blue}{0.095} & \textcolor{red}{41.34} & \textcolor{red}{0.984} & \textcolor{red}{0.012} \\

& 3DGS      & 11.78 & 0.353 & 0.555 & 9.72 	& 0.473 & 0.551 & 10.94 & 0.380 & 0.550 & 10.58 & 0.316 & 0.600  \\

& Ours      & \textcolor{red}{39.64} & \textcolor{red}{0.970} & \textcolor{red}{0.013} & \textcolor{red}{43.16} & \textcolor{red}{0.992} & \textcolor{red}{0.003} & \textcolor{red}{32.84} & \textcolor{red}{0.922} & \textcolor{red}{0.034} & \textcolor{blue}{37.66} & 0.972 & 0.023 \\ 

\midrule[1pt]

\multirow{4}{*}{LDR-NE} 
& NeRF \tnote{*}     & - & - & - & - & - & - & - & - & - & - & - & - \\ 
& HDR-P   \tnote{+} & 6.64 &	0.349 	&0.662&10.70&	0.636 &	0.551 &10.27 &	0.403 &	0.862 &11.35 &	0.439 	&0.669  \\ 
& NeRF-W  \tnote{*}  & 26.98 & 0.881 & \textcolor{blue}{0.066} & 32.67 & 0.976 & \textcolor{blue}{0.022} & 26.96 & 0.815 & 0.157 & 32.53 & 0.972 & 0.019 \\ 
& HDR-NeRF  & \textcolor{blue}{32.49} & \textcolor{blue}{0.942} & 0.076 & \textcolor{blue}{38.48} & \textcolor{blue}{0.987} & 0.031 & \textcolor{red}{32.50} & \textcolor{blue}{0.909} & \textcolor{blue}{0.091} & \textcolor{red}{41.71} & \textcolor{red}{0.988} & \textcolor{red}{0.009} \\
& 3DGS      & 12.94 & 0.464 & 0.420 & 10.83 & 0.521 & 0.500 & 12.05 & 0.422 & 0.484 & 11.05 & 0.322 & 0.594 \\
& Ours      & \textcolor{red}{39.61} & \textcolor{red}{0.973} & \textcolor{red}{0.014} & \textcolor{red}{42.56} & \textcolor{red}{0.991} & \textcolor{red}{0.005} & \textcolor{blue}{32.49} & \textcolor{red}{0.923} & \textcolor{red}{0.037} & \textcolor{blue}{38.01} & \textcolor{blue}{0.980} & \textcolor{blue}{0.013} \\

\midrule[2pt]
\multirow{2}{*}{} & \multirow{2}{*}{} & \multicolumn{3}{c}{Dog} & \multicolumn{3}{c}{Desk} & \multicolumn{3}{c}{Sofa} & \multicolumn{3}{c}{Sponza}  \\
    &   & \multicolumn{1}{c}{PSNR}\(\uparrow\) & \multicolumn{1}{c}{SSIM} \(\uparrow\) & \multicolumn{1}{c}{LPIPS}\(\downarrow \)
        & \multicolumn{1}{c}{PSNR}\(\uparrow\)& \multicolumn{1}{c}{SSIM} \(\uparrow\) & \multicolumn{1}{c}{LPIPS}\(\downarrow \)
        & \multicolumn{1}{c}{PSNR}\(\uparrow\) & \multicolumn{1}{c}{SSIM} \(\uparrow\) & \multicolumn{1}{c}{LPIPS}\(\downarrow \)
        & \multicolumn{1}{c}{PSNR}\(\uparrow\)& \multicolumn{1}{c}{SSIM} \(\uparrow\) & \multicolumn{1}{c}{LPIPS}\(\downarrow \)
        \\ 
\midrule[1pt]
\multirow{4}{*}{LDR-OE} 
& NeRF  \tnote{*}    & 13.69 & 0.619 & 0.279 & 15.29 & 0.645 & 0.249 & 15.06 & 0.718 & 0.229 & 16.39 & 0.664 & 0.219 \\ 
& HDR-P \tnote{ +} & 10.48 &	0.395 	&0.737& 8.69 	&0.455 &	0.660& 10.34& 	0.444 &	0.783& 10.33 &	0.358 &	0.778 \\
& NeRF-W \tnote{*}   & 31.01 & \textcolor{blue}{0.967} & \textcolor{blue}{0.022} & 30.21 & 0.958 & 0.030 & 30.76 & 0.955 & 0.029 & 24.50 & 0.908 & 0.037 \\ 
& HDR-NeRF   & \textcolor{blue}{34.46} & 0.948 & 0.040 & \textcolor{blue}{37.62} & \textcolor{blue}{0.965} & \textcolor{blue}{0.026}  & \textcolor{blue}{37.88} & \textcolor{blue}{0.973} & \textcolor{blue}{0.017} & \textcolor{blue}{34.85} & \textcolor{blue}{0.959} & \textcolor{blue}{0.030} \\
& 3DGS      & 8.51 	& 0.193 & 0.651 & 11.54 & 0.429 & 0.484 & 10.37 & 0.370 & 0.562 & 16.46 & 0.645 & 0.317 \\
& Ours      & \textcolor{red}{40.80} & \textcolor{red}{0.990} & \textcolor{red}{0.005} & \textcolor{red}{40.43} & \textcolor{red}{0.981} & \textcolor{red}{0.007} & \textcolor{red}{40.32} & \textcolor{red}{0.986} & \textcolor{red}{0.007} & \textcolor{red}{38.42} & \textcolor{red}{0.980} & \textcolor{red}{0.008} \\ 
\midrule[1pt]

\multirow{4}{*}{LDR-NE} 
& NeRF   \tnote{*}   & - & - & - & - & - & - & - & - & - & - & - & - \\ 
& HDR-P \tnote{ +} & 10.48 	&0.395 	&0.737& 8.62	&0.455 &	0.660& 10.32 	&0.444 &	0.789 &10.33 &	0.358 &	0.778 \\
& NeRF-W \tnote{*}   & 30.41 & \textcolor{blue}{0.964} & \textcolor{blue}{0.026} & 29.60 & \textcolor{blue}{0.950} & \textcolor{blue}{0.034} & 30.31 & 0.952 & \textcolor{blue}{0.031} & 24.32 & 0.904 & \textcolor{blue}{0.042} \\
& HDR-NeRF  & \textcolor{blue}{31.66} & 0.939 & 0.088 & \textcolor{blue}{31.86} & 0.937 & 0.092  & \textcolor{blue}{36.05} & \textcolor{blue}{0.962} & 0.036 & \textcolor{blue}{30.72} & \textcolor{blue}{0.924} & 0.092 \\
& 3DGS      & 9.09 	& 0.197 & 0.648 & 11.23 & 0.414 & 0.502  & 11.89 & 0.462 & 0.465 & 16.71 & 0.649 & 0.298 \\
& Ours      & \textcolor{red}{40.71} & \textcolor{red}{0.990} & \textcolor{red}{0.005} & \textcolor{red}{39.83} & \textcolor{red}{0.979}	& \textcolor{red}{0.008} & \textcolor{red}{39.07} & \textcolor{red}{0.983} & \textcolor{red}{0.009} & \textcolor{red}{38.47} & \textcolor{red}{0.979} & \textcolor{red}{0.009} \\
\bottomrule[2pt]
\end{tabular}
}
\begin{tablenotes}
        \footnotesize
        \item [ \(+\)] HDR-Plenoxel\cite{jun2022hdr}. We directly utilize the grid parameters provided by them for training, which may not \\be suitable for HDRNeRF datasets.
        \item [ \(*\)] The results are derived from HDRNeRF\cite{huang2022hdr}.
\end{tablenotes}

\end{threeparttable}

\end{table}

\begin{table}[] 
\centering
\caption{ LDR image quality assessment, LDR-OE represents the use of LDR images with exposures \(\{t_1,t_3,t_5\}\) as the training set, while LDR-NE represents the use of LDR images with exposures \(\{t_2,t_4\}\) as the training set. The exposure times of the test dataset are \( \{ t_1, t_2, t_3, t_4, t_5 \} \). The \textcolor{red}{best} and the \textcolor{blue}{second best} results are denoted by red and blue.}
\label{tab:all real ldr}
    
\begin{threeparttable}
\resizebox{\textwidth}{!} 
{
\begin{tabular}{c|c|ccccccccccccc}
\toprule[2pt]
\multirow{2}{*}{} & \multirow{2}{*}{} & \multicolumn{3}{c}{box} & \multicolumn{3}{c}{computer} & \multicolumn{3}{c}{flower} & \multicolumn{3}{c}{luckycat}\\
    &   & \multicolumn{1}{c}{PSNR}\(\uparrow\) & \multicolumn{1}{c}{SSIM} \(\uparrow\) & \multicolumn{1}{c}{LPIPS}\(\downarrow \)
        & \multicolumn{1}{c}{PSNR}\(\uparrow\) & \multicolumn{1}{c}{SSIM} \(\uparrow\) & \multicolumn{1}{c}{LPIPS}\(\downarrow \)
        & \multicolumn{1}{c}{PSNR}\(\uparrow\) & \multicolumn{1}{c}{SSIM} \(\uparrow\) & \multicolumn{1}{c}{LPIPS}\(\downarrow \)
        & \multicolumn{1}{c}{PSNR}\(\uparrow\) & \multicolumn{1}{c}{SSIM} \(\uparrow\) & \multicolumn{1}{c}{LPIPS}\(\downarrow \)
        \\ 
\midrule[1pt]

\multirow{4}{*}{LDR-OE} 
& NeRF  \tnote{*}    & 17.06 & 0.770 & 0.233 & 14.68 & 0.697 & 0.281 & 14.60 & 0.504 & 0.524 & 13.67 & 0.706 & 0.262\\
& HDR-P \tnote{ \(+\)} & 9.47 &	0.587& 	0.693 &8.91 &	0.469 &	0.697 & 9.30 &	0.599 	&0.711& 9.39 &	0.610 &	0.696  \\ 
& NeRF-W \tnote{*}   & 29.21 & 0.927 & 0.097 & 28.91 & 0.919 & 0.112 & 26.23 & 0.933 & 0.094 & 30.00 & 0.927 & 0.076\\
& HDR-NeRF  & \textcolor{blue}{31.48} & \textcolor{blue}{0.957} & \textcolor{blue}{0.062} & \textcolor{blue}{32.55} & \textcolor{blue}{0.945} & \textcolor{blue}{0.079} & \textcolor{blue}{31.01} & \textcolor{blue}{0.952} & \textcolor{blue}{0.061} & \textcolor{blue}{32.47} & \textcolor{blue}{0.945} & \textcolor{blue}{0.063} \\
& 3DGS      & 12.25 & 0.628 & 0.360 & 11.15 & 0.511 & 0.367 & 12.15 & 0.651 & 0.342 & 11.91 & 0.654 & 0.352 \\
& Ours      & \textcolor{red}{33.65} & \textcolor{red}{0.972} & \textcolor{red}{0.021}  & \textcolor{red}{32.79} & \textcolor{red}{0.961} & \textcolor{red}{0.026} & \textcolor{red}{31.28} & \textcolor{red}{0.964} & \textcolor{red}{0.028} & \textcolor{red}{35.64} & \textcolor{red}{0.972} & \textcolor{red}{0.017} \\
\midrule[1pt]

\multirow{4}{*}{LDR-NE} 
& NeRF   \tnote{*}   & - & - & -  & - & - & - & - & - & - & - & - & - \\ 
& HDR-P \tnote{ \(+\)}  &9.41& 	0.588 &	0.694& 8.88 &	0.471 	&0.700& 9.25 &	0.600 &	0.710 &9.33 &	0.613 &	0.696  \\ 
& NeRF-W \tnote{*}   & 29.59 & 0.923 & 0.104 & 27.54 & 0.892 & 0.136 & 26.84 & 0.939 & 0.078 & \textcolor{blue}{30.78} & \textcolor{blue}{0.940} & \textcolor{blue}{0.058} \\
& HDR-NeRF  & \textcolor{blue}{30.21} & \textcolor{blue}{0.955} & \textcolor{blue}{0.073} & \textcolor{red}{32.48} & \textcolor{red}{0.945} & \textcolor{blue}{0.083} & \textcolor{blue}{30.15} & \textcolor{blue}{0.948} & \textcolor{blue}{0.063} & 28.90 & 0.936 & 0.094 \\
& 3DGS      & 12.76 & 0.694 & 0.342 & 13.40 & 0.594 & 0.353 & 12.93 & 0.717 & 0.332 & 12.85 & 0.724 & 0.330 \\
& Ours      & \textcolor{red}{32.94} & \textcolor{red}{0.971} & \textcolor{red}{0.020} & \textcolor{blue}{32.30} & \textcolor{blue}{0.932} & \textcolor{red}{0.056} & \textcolor{red}{30.79} & \textcolor{red}{0.964} & \textcolor{red}{0.022} & \textcolor{red}{35.05} & \textcolor{red}{0.972} & \textcolor{red}{0.019} \\
\bottomrule[2pt]
\end{tabular}

}
\begin{tablenotes}
        \footnotesize
        \item [ \(+\)] HDR-Plenoxel\cite{jun2022hdr}. We directly utilize the grid parameters provided by them for training, which may not \\be suitable for HDRNeRF datasets.
        \item [ \(*\)] The results are derived from HDRNeRF\cite{huang2022hdr}.
\end{tablenotes}
      
\end{threeparttable}
\end{table}

\begin{table}[]  
\centering
\caption{ HDR image quality assessment, LDR-OE represents the use of LDR images with exposures \(\{t_1,t_3,t_5\}\) as the training set, while LDR-NE represents the use of LDR images with exposures \(\{t_2,t_4\}\) as the training set. The best results are denoted by red.} 
\label{tab:all hdr}
    
\resizebox{\textwidth}{!}
{
\begin{tabular}{c|c|cccccccccccc}
\toprule[2pt]
\multirow{2}{*}{ } & \multirow{2}{*}{ }  & \multicolumn{3}{c}{Bathroom} & \multicolumn{3}{c}{Bear} & \multicolumn{3}{c}{Chair} \\
    &  & \multicolumn{1}{c}{HDR VDP-Q}\(\uparrow\) & \multicolumn{1}{c}{PUPSNR} \(\uparrow\)  & \multicolumn{1}{c}{PUSSIM}\(\uparrow\)
        & \multicolumn{1}{c}{HDR VDP-Q}\(\uparrow\) & \multicolumn{1}{c}{PUPSNR}\(\uparrow\) & \multicolumn{1}{c}{PUSSIM}\(\uparrow\)
        & \multicolumn{1}{c}{HDR VDP-Q}\(\uparrow\) & \multicolumn{1}{c}{PUPSNR} \(\uparrow\) & \multicolumn{1}{c}{PUSSIM}\(\uparrow\)
        \\ 
\midrule[1 pt]
\multirow{2}{*}{LDR-OE} 
& HDR-NeRF  & 9.50 & 22.42 & 0.88 & 9.94 & \textcolor{red}{37.65} & \textcolor{red}{0.99} & 9.11 & 17.91 & 0.40    \\
& Ours      & \textcolor{red}{9.85} & \textcolor{red}{26.92} & \textcolor{red}{0.92} & \textcolor{red}{9.95} & 28.78 & 0.97 & \textcolor{red}{9.40} & \textcolor{red}{18.37} & \textcolor{red}{0.49}    \\ 
\midrule[1 pt]
\multirow{2}{*}{LDR-NE} 
& HDR-NeRF  & 1.85 & 7.57  & 0.02 & -0.22 & 5.64 & 0.01 & 5.05 & 14.63 & 0.21    \\
& Ours      & \textcolor{red}{9.91} & \textcolor{red}{27.92} & \textcolor{red}{0.94} & \textcolor{red}{9.94} & \textcolor{red}{19.18} & \textcolor{red}{0.83} & \textcolor{red}{9.37} & \textcolor{red}{15.48} & \textcolor{red}{0.22}    \\

\midrule[2 pt]
\multirow{2}{*}{} & \multirow{2}{*}{} & \multicolumn{3}{c}{Diningroom} & \multicolumn{3}{c}{Dog} & \multicolumn{3}{c}{Desk} \\
    &   & \multicolumn{1}{c}{HDR VDP-Q}\(\uparrow\) & \multicolumn{1}{c}{PUPSNR} \(\uparrow\) & \multicolumn{1}{c}{PUSSIM}\(\uparrow\)
        & \multicolumn{1}{c}{HDR VDP-Q}\(\uparrow\) & \multicolumn{1}{c}{PUPSNR} \(\uparrow\) & \multicolumn{1}{c}{PUSSIM}\(\uparrow\) 
        & \multicolumn{1}{c}{HDR VDP-Q}\(\uparrow\)& \multicolumn{1}{c}{PUPSNR} \(\uparrow\) & \multicolumn{1}{c}{PUSSIM}\(\uparrow\) 
        \\ 
\midrule[1 pt]
\multirow{2}{*}{LDR-OE} 
& HDR-NeRF  & \textcolor{red}{9.94} & \textcolor{red}{35.64} & \textcolor{red}{1.00} & 9.57 & 18.64 & 0.94 & 9.65 & \textcolor{red}{33.60} & \textcolor{red}{0.92}    \\
& Ours      & 9.83 & 25.13 & 0.93 & \textcolor{red}{9.87} & \textcolor{red}{26.62} & \textcolor{red}{0.97} & \textcolor{red}{9.73} & 31.77 & 0.87   \\ 
\midrule[1 pt]
\multirow{2}{*}{LDR-NE} 
& HDR-NeRF  & \textcolor{red}{9.87} & \textcolor{red}{28.60}  & 0.86 & 9.46 & 11.69 & -0.12 & -0.82 & 19.94 & 0.36    \\
& Ours      & 9.25 & 24.27 & \textcolor{red}{0.96} & \textcolor{red}{9.79} & \textcolor{red}{25.14} & \textcolor{red}{0.94} & \textcolor{red}{9.63} & \textcolor{red}{32.14} & \textcolor{red}{0.91}    \\

\midrule[2 pt]
\multirow{2}{*}{} & \multirow{2}{*}{} & \multicolumn{3}{c}{sofa} & \multicolumn{3}{c}{sponza} & \multicolumn{3}{c}{avg.}\\
    &   & \multicolumn{1}{c}{HDR VDP-Q}\(\uparrow\) & \multicolumn{1}{c}{PUPSNR} \(\uparrow\) & \multicolumn{1}{c}{PUSSIM}\(\uparrow\)
        & \multicolumn{1}{c}{HDR VDP-Q}\(\uparrow\) & \multicolumn{1}{c}{PUPSNR} \(\uparrow\) & \multicolumn{1}{c}{PUSSIM}\(\uparrow\) 
        & \multicolumn{1}{c}{HDR VDP-Q}\(\uparrow\)& \multicolumn{1}{c}{PUPSNR} \(\uparrow\) & \multicolumn{1}{c}{PUSSIM}\(\uparrow\) 
        \\ 
\midrule[1 pt]
\multirow{2}{*}{LDR-OE} 
& HDR-NeRF  & \textcolor{red}{9.75} & \textcolor{red}{31.24} & \textcolor{red}{0.99} & 2.14 & 6.11 & 0.00 & 8.70 & \textcolor{red}{25.40} & 0.76 \\
& Ours      & 9.52 & 25.37 & 0.88 & \textcolor{red}{9.89} & \textcolor{red}{17.37} & \textcolor{red}{0.83} & \textcolor{red}{9.75} & 25.04 & \textcolor{red}{0.86} \\ 
\midrule[1 pt]
\multirow{2}{*}{LDR-NE} 
& HDR-NeRF  & \textcolor{red}{9.70} & \textcolor{red}{29.24} & \textcolor{red}{0.99} & 0.42 & 4.82 & 0.00  & 4.41 & 15.27 & 0.29 \\
& Ours      & 9.62 & 17.63 & 0.87 & \textcolor{red}{9.70} & \textcolor{red}{16.65} & \textcolor{red}{0.82} & \textcolor{red}{9.65} & \textcolor{red}{22.30} & \textcolor{red}{0.81} \\

\bottomrule[2 pt]
\end{tabular}
}
\end{table}

\begin{figure}
\centering
\subfloat{
		\includegraphics[scale=0.80]{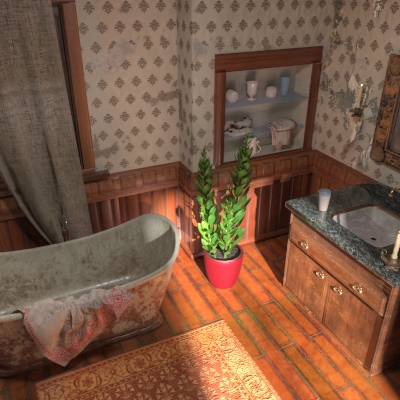}}
\subfloat{
		\includegraphics[scale=0.8]{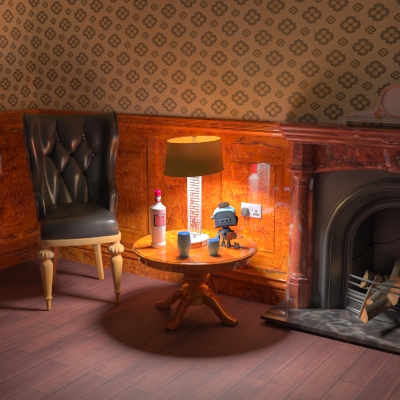}}
\subfloat{
		\includegraphics[scale=0.8]{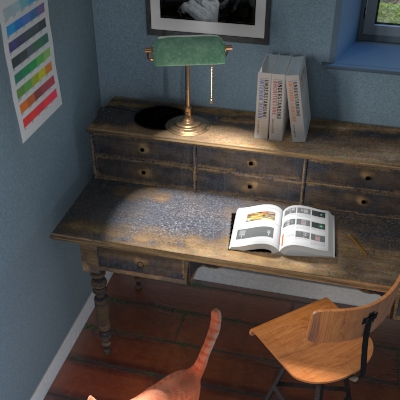}}
\subfloat{
		\includegraphics[scale=0.8]{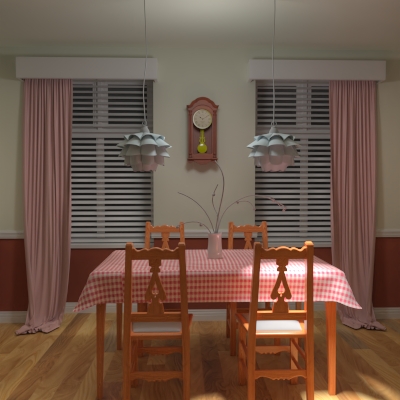}}
\subfloat{
		\includegraphics[scale=0.8]{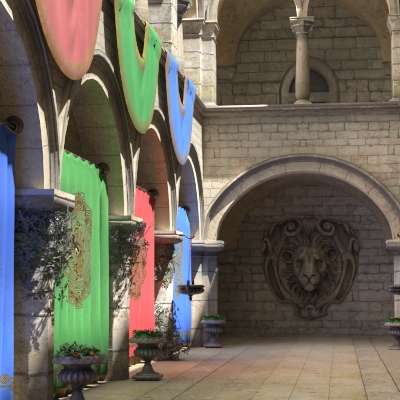}}

\subfloat{
		\includegraphics[scale=0.8]{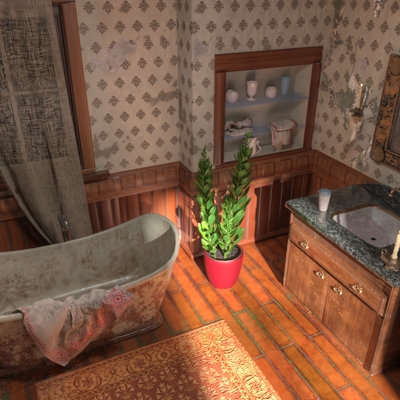}}
\subfloat{
		\includegraphics[scale=0.8]{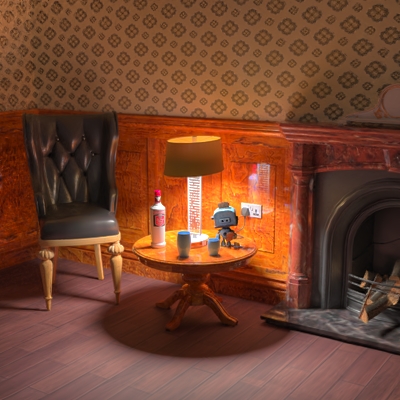}}
\subfloat{
		\includegraphics[scale=0.8]{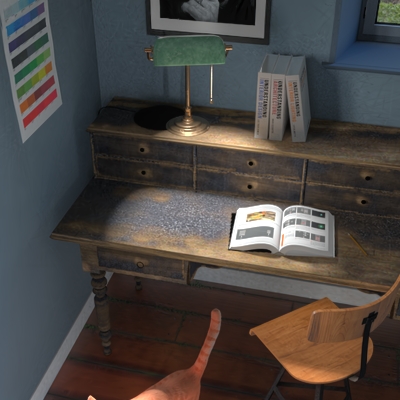}}
\subfloat{
		\includegraphics[scale=0.8]{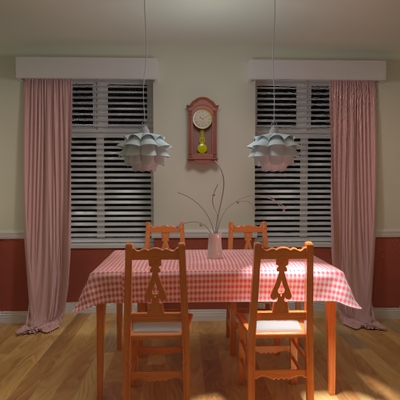}}
\subfloat{
		\includegraphics[scale=0.8]{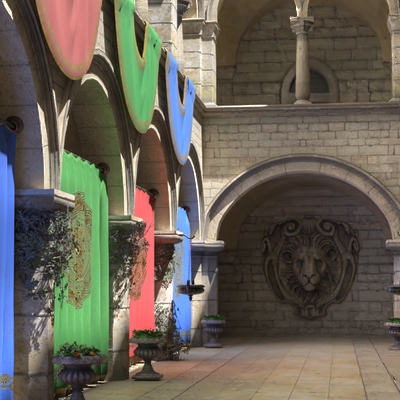}}

\setcounter{subfigure}{0}  

\subfloat[bathroom]{
		\includegraphics[scale=0.8]{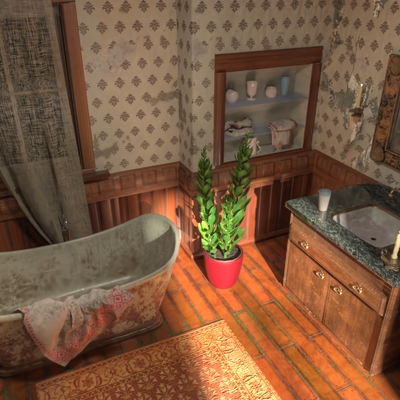}}
\subfloat[chair]{
		\includegraphics[scale=0.8]{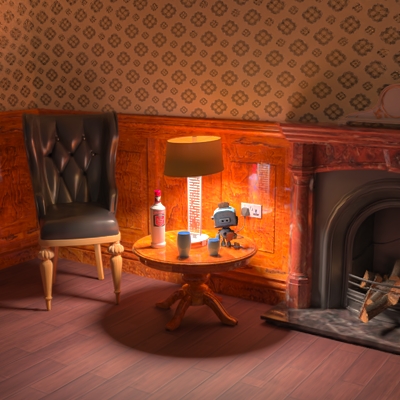}}
\subfloat[desk]{
		\includegraphics[scale=0.8]{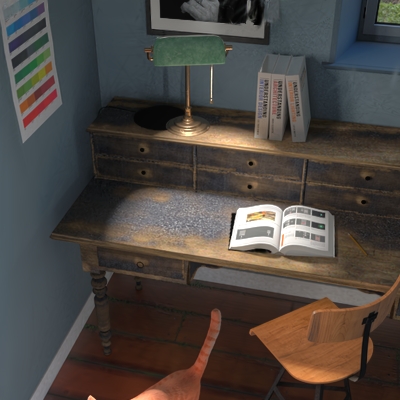}}
\subfloat[diningroom]{
		\includegraphics[scale=0.8]{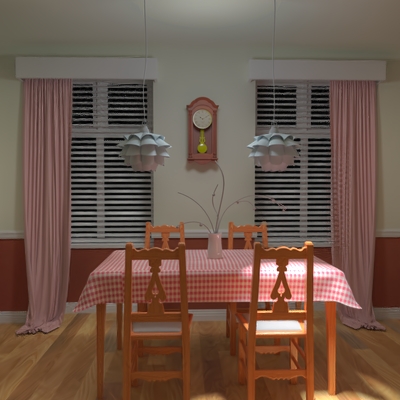}}
\subfloat[sponza]{
		\includegraphics[scale=0.8]{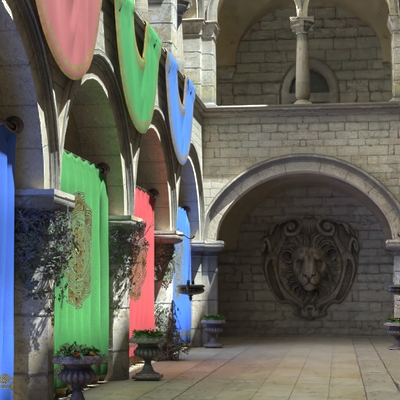}}

\caption{HDR image quality comparison: the first row shows the ground truth (GT). The second and third rows respectively show HDR images rendered by our method with \(\mathcal L_u\) and without \(\mathcal L_u\).}

\label{fig:hdr ablation}
\end{figure}

\end{document}